\pdfoutput=1
\documentclass{aifrontiers}

\usepackage{aifrontiers}
\usepackage[utf8]{inputenc}
\usepackage[T1]{fontenc}
\usepackage{booktabs}
\usepackage{tabularx}
\usepackage{makecell}
\usepackage{multirow}
\usepackage{array}
\usepackage{longtable}
\usepackage{etoc}
\usepackage{colortbl}
\usepackage{caption}
\usepackage{subcaption}
\usepackage{multicol}
\usepackage{float}

\usepackage{enumitem}
\usepackage{xspace}
\usepackage{listings}
\usepackage{amsmath}
\usepackage{xcolor}
\usepackage[export]{adjustbox}
\usepackage{pifont}

\definecolor{apbslate}{HTML}{5A6B7C}
\definecolor{fgIndigo}{HTML}{3B5BA5}
\definecolor{fgTeal}{HTML}{2E8B84}
\definecolor{fgAmber}{HTML}{D89B3C}
\definecolor{fgRose}{HTML}{B56B80}
\definecolor{fgPlum}{HTML}{7A5B9B}
\definecolor{fgOlive}{HTML}{6E8B3D}
\definecolor{fgSlate}{HTML}{3F8A8C}
\definecolor{fgClay}{HTML}{B4704A}
\definecolor{fgMoss}{HTML}{57806B}

\usetikzlibrary{arrows.meta, positioning, calc, patterns, patterns.meta, fit}
\usepackage{pgfplots}
\pgfplotsset{compat=1.18}

\titlespacing*{\section}      {0pt}{1.6ex plus .2ex}{0.9ex}
\titlespacing*{\subsection}   {0pt}{1.4ex plus .2ex}{0.7ex}
\titlespacing*{\subsubsection}{0pt}{1.2ex plus .2ex}{0.6ex}

\titleformat{\paragraph}[runin]{\normalfont\bfseries}{}{0pt}{}[.]
\titlespacing*{\paragraph}{0pt}{1.1ex plus .2ex}{0.6em}

\renewcommand{\arraystretch}{1.15}

\newcommand{\env}[1]{#1}

\definecolor{apbblue}{HTML}{2B6CB0}

\definecolor{openaiblack}{HTML}{000000}
\definecolor{anthropicorange}{HTML}{D97757}
\definecolor{googleblue}{HTML}{4285F4}
\definecolor{qwenpurple}{HTML}{615CED}
\definecolor{deepseekblue}{HTML}{4D6BFE}

\definecolor{surveyc}{HTML}{2B6CB0}
\definecolor{chatc}{HTML}{2F855A}
\definecolor{webc}{HTML}{B7791F}
\definecolor{appc}{HTML}{9B2C2C}

\newcommand{\vlogo}[1]{%
  \raisebox{-0.14em}{\includegraphics[height=0.92em]{figures/logos/#1}}%
}
\newcommand{\iconopenai}{\vlogo{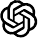}}
\newcommand{\iconclaude}{\vlogo{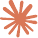}}
\newcommand{\icongemini}{\vlogo{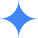}}
\newcommand{\iconqwen}{\vlogo{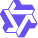}}
\newcommand{\icondeepseek}{\vlogo{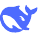}}

\newcommand{\iconkimi}{\vlogo{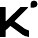}}
\newcommand{\iconglm}{\vlogo{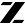}}
\newcommand{\icongrok}{\vlogo{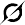}}

\title{AgentPersonaBench: Benchmarking\\Persona-Driven User Simulation}
\shorttitle{AgentPersonaBench}

\newcommand{\organizerauthors}{%
Jintao Huang$^{2}$,
Yifan Wang$^{3}$}%

\newcommand{\contributorauthors}{%
Hongyu Shen$^{4}$,
Yi Daniel Lu$^{5}$,
Shirley Huang$^{5}$,
Minsik Oh$^{6}$,
Yewen Wang$^{7}$,
Muhammad Ahmed Mohsin$^{6}$,
Zhen Xu$^{8}$,
Yilan Fan$^{9}$,
Zichen Yuan$^{10}$,
Ahsan Bilal$^{11}$,
Zibu Wei$^{7}$,
Sankalp Jajee$^{12}$,
Henry Gagnier$^{13}$,
Saksham Kapoor$^{14}$,
Jicheng Wang$^{15}$,
Qianfeng Wen$^{16}$,
Yixuan He$^{17}$,
Steven Dillmann$^{6}$,
Jiashu He$^{18}$,
Yucheng Lu$^{9}$,
Linqiang Guo$^{19}$,
Danyang Zhang$^{20}$,
Shi Bo$^{21}$,
Raunak Mondal$^{22}$,
Haixiang Tang$^{4}$,
Weihang Xiao$^{23}$,
Allen Nie$^{6}$,
Jing Tang$^{21}$,
Yueying Li$^{24}$,
Yifan Simon Liu$^{16}$,
Jianheng Hou$^{25}$,
Dianzhuo Wang$^{5}$,
Qianyu Zhu$^{24}$,
Zhixu Silvia Tao$^{26}$,
Zhejian Peng$^{4}$,
Zihan Wang$^{27}$,
Ishan Gupta$^{28}$,
Jinxuan Fan$^{29}$,
Wanting Jiang$^{30}$,
Shushu Liang$^{5}$}%

\newcommand{\advisoryauthors}{%
Chenxi Qiu$^{31}$,
Yijun Wang$^{5}$,
Xiaomin Li$^{1,5}$,
Yuexing Hao$^{1,24}$
}

\newcommand{\authoraffiliations}{%
\small
\begin{multicols}{2}
\begin{enumerate}[leftmargin=1.6em,itemsep=0.15em,topsep=0pt]
  \item MatrAIx
  \item The Ohio State University
  \item Purdue University
  \item University of Illinois Urbana-Champaign
  \item Harvard University
  \item Stanford University
  \item University of California, Los Angeles
  \item Columbia University
  \item New York University
  \item Washington University in St.\ Louis
  \item University of Oklahoma
  \item Medical University of South Carolina
  \item University of Rochester
  \item University of Maryland, College Park
  \item University of California, Davis
  \item University of Toronto
  \item Arizona State University
  \item University of Pennsylvania
  \item Mila, Quebec AI Institute
  \item San Jose State University
  \item Boston University
  \item Carnegie Mellon University
  \item Cornell University
  \item Massachusetts Institute of Technology
  \item University of Southern California
  \item Princeton University
  \item Abaka AI
  \item University of California, San Diego
  \item University of Texas at Austin
  \item Amherst College
  \item Harvard Medical School
\end{enumerate}
\end{multicols}
}

\author{
{\normalsize \organizerauthors, \contributorauthors, \advisoryauthors}
}
\date{}

\renewcommand{\AIFTitleAuthorGap}{0.8em}
\renewcommand{\AIFAbstractTopGap}{6pt}
\renewcommand{\AIFAbstractBottomGap}{24pt}
\renewcommand{\AIFAbstractVerticalPadding}{5pt}

\begin{document}
\etocdepthtag.toc{mtmain}
\etocsettagdepth{mtmain}{none}
\etocsettagdepth{mtappendix}{subsection}

\begin{abstract}
We introduce \textbf{AgentPersonaBench (APB)}\footnotemark, a benchmark evaluating whether
persona conditioning faithfully steers downstream agent behavior. While language
models are increasingly deployed for \emph{persona-driven user simulation}, existing
benchmarks primarily evaluate conversational styling or self-reports rather than authentic
behavioral fidelity. APB evaluates latent persona adherence one trait at a time,
embedding each target trait within a complete synthetic profile without explicitly
naming the trait or disclosing the test. Ground-truth adherence is verified strictly
from observable actions across four interaction surfaces of increasing realism:
\env{survey}, \env{chat}, \env{web} (interactive web environments), and \env{app} (desktop
software environments). APB comprises \textbf{2{,}460 tasks} spanning
867 traits, verified through automated audits and expert review.
Our evaluation of 20 frontier model arms demonstrates that high-fidelity user
simulation is already attainable: leading models achieve up to 84.7\% full-pass
adherence under unprompted conditions. At the same time, APB identifies clear
behavioral boundaries: adherence drops across interaction modalities (only 37.9--64.3\% pass
all four surfaces), multi-attribute demands degrade retention, and competing model families
exhibit pronounced behavioral divergence.
\end{abstract}
\maketitle
\footnotetext{The benchmark dataset, environments, and evaluation code are publicly available at \url{https://github.com/MatrAIx-ai/AgentPersonaBench}.}

\noindent\begin{minipage}{1\textwidth}
\centering
\captionsetup{font=small, hypcap=false, skip=4pt}
\includegraphics[width=\textwidth]{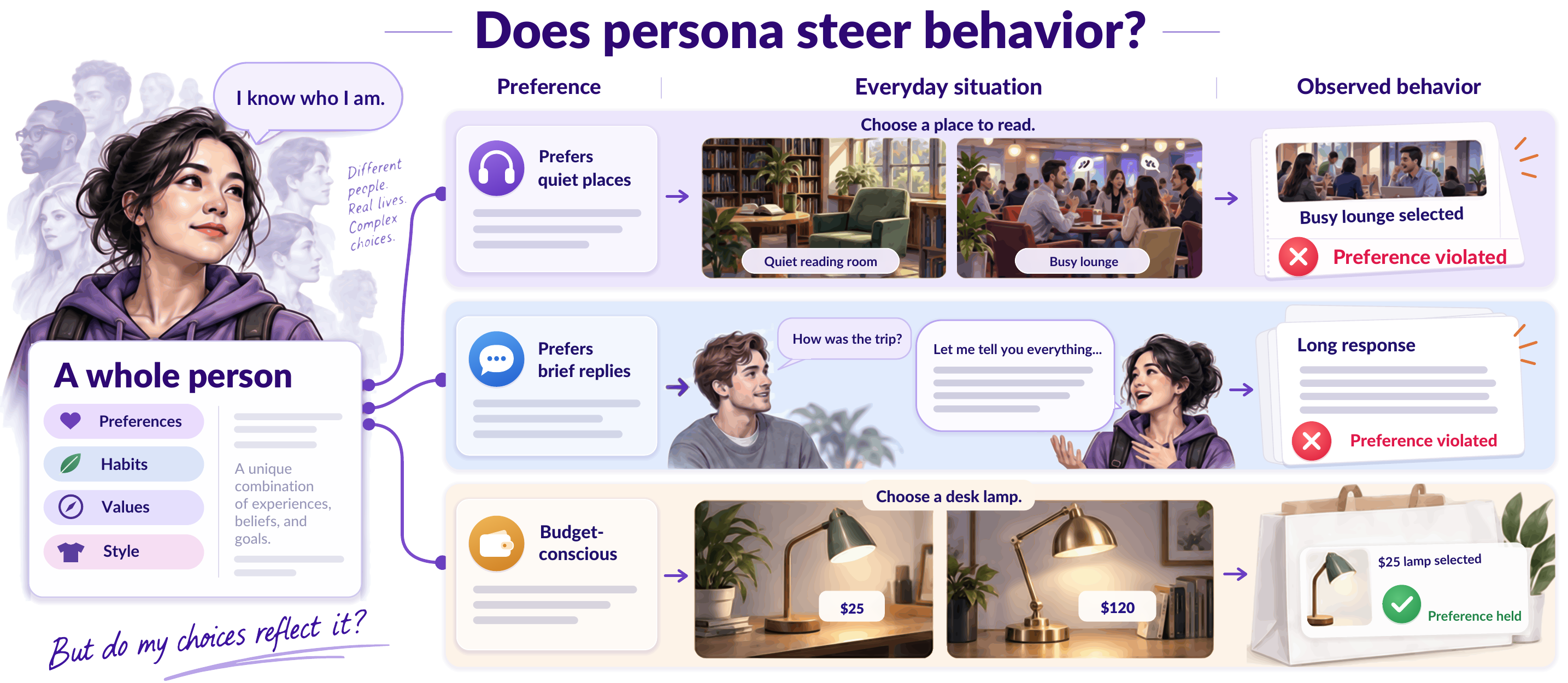}
\captionof{figure}{\textbf{Does persona steer behavior?}
A rich persona may define an agent's assigned profile without consistently
governing its downstream choices. Everyday situations reveal whether latent persona
traits are faithfully maintained or violated. Personas, situations, and outcomes
are illustrative.}
\label{fig:introduction}
\end{minipage}\par

\section{Introduction}
\label{sec:intro}

Deployed language-model systems increasingly embody personal identities. Recommender agents reflect consumer tastes, social science experiments deploy synthetic cohorts to simulate human respondents, autonomous agents stand in for users during software testing, and personal assistants operate under persistent user profiles. In each setting, a system prompt establishes an identity, and the model is entrusted to sustain it across diverse tasks without explicit reminders. We define this paradigm as \emph{persona-driven user simulation}: a persona specifies an individual's latent attributes, an agent acts them out, and consequential downstream decisions are made based on the agent's observable behavior. Yet existing evaluations focus almost exclusively on conversational role-play: maintaining stylistic tone, avoiding breaking frame, and recalling biographical trivia. These criteria evaluate only the conversational surface. When an agent acts on a user's behalf, the decisive question is behavioral: does a latent trait observably govern what the agent decides in an everyday situation where that trait is never mentioned? As illustrated in Figure~\ref{fig:introduction}, a rich persona may fully describe who an agent is without consistently steering what it does. A vegan persona ordering a meal must not place meat in the shopping cart, and a risk-averse persona rebalancing a portfolio must avoid high-leverage assets. If the trait genuinely guides the agent, it directly dictates the physical choice; if the model merely generates persona-flavored prose, the behavioral constraint is lost.

To evaluate authentic behavioral steerability, we introduce \textbf{AgentPersonaBench (APB)}, a benchmark evaluating whether latent persona conditioning faithfully guides downstream agent behavior across everyday digital tasks. APB comprises 2,460 tasks spanning 867 traits. Rather than testing traits in isolation with minimal stub prompts, APB embeds each target trait into a complete individual profile sampled from a one-million persona corpus. Because the profile spans hundreds of realistic background attributes, the tested trait must actively compete for attention, faithfully reproducing the distracting conditions of production deployments.

To evaluate steerability under varied interaction modalities, tasks are constructed across four interaction surfaces of increasing environmental realism (\S\ref{sec:surfaces}): \env{survey} (questionnaires with randomized options), \env{chat} (dialogues with scripted off-persona pressure), \env{web} (interactive web environments evaluated from browser state and interface choices), and \env{app} (desktop applications evaluated through visual interaction and application state). Across all surfaces, tasks are constructed under strict latent conditioning: the target trait is never named, the evaluation is never announced, and ground truth is strictly isolated from the agent's view. Every task is audited to eliminate guessing shortcuts, position bias, and superficial prompt leaks before admission (\S\ref{sec:verification}).

APB evaluates model performance strictly from observable environment artifacts rather than conversational self-reports or post-hoc justifications. Approximately four-fifths (78.5\%) of all checks are deterministic code predicates over structured artifacts (selected option IDs, DOM state, and application files), reserving standardized rubric evaluation solely for open-ended dialogue turns. Across 20 frontier model arms from eight providers, our evaluation shows that leading models can achieve up to 84.7\% full-pass adherence under unprompted conditions, while identifying cross-surface fragility and multi-attribute co-exposure as primary open challenges.

This paper provides three primary contributions:
\begin{itemize}[leftmargin=1.4em, itemsep=1pt, topsep=1pt]
  \item \textbf{A Latent Behavioral Measurement Protocol}: We formalize an evaluation methodology that assesses persona adherence exclusively through unprompted, verified actions rather than conversational self-reports or explicit instructions, enforcing strict leak prevention and audit verification.
  \item \textbf{A 2,460-Task Multi-Surface Benchmark}: We construct and release the first comprehensive benchmark evaluating latent persona adherence across four distinct interaction surfaces (including interactive web environments and desktop graphical interfaces) spanning 867 traits.
  \item \textbf{Empirical Findings on Frontier Models}: We evaluate 20 model arms across eight providers. Leading models (including \texttt{gemini-3-8-flash} at 84.7\%, \texttt{opus-5-5} at 82.9\%, and \texttt{gpt-6-astra} at 82.1\%) show that reliable persona simulation is achievable, while our diagnostics identify cross-surface fragility and multi-attribute co-exposure as key challenges.
\end{itemize}

\section{Related Work}
\label{sec:related}

\paragraph{User simulation and persona agents.}
Language models are widely evaluated as persona agents, either through conversational
role-play benchmarks assessing voice consistency, frame retention, and biographical
recall \citep{zhang2018personachat, wang2023rolellm, tu2024charactereval,
samuel2024personagym, zhou2024sotopia, tseng2024two}, or through synthetic cohorts
simulating demographic populations and human subjects in social science
\citep{argyle2023out, santurkar2023whose, horton2023large, park2023generative,
park2024generative1000, ge2024scaling, yao2024tau}. However, conversational
benchmarks evaluate whether generated text ``sounds like'' a character; an agent
writing eloquent prose while selecting meat for a vegan passes text-based role-play
rubrics but fails in practice. Meanwhile, demographic simulations evaluate aggregate
statistical alignment, where individual errors can cancel out across cohorts. In
contrast, \textsc{APB} measures \emph{individual behavioral fidelity under
deterministic ground truth}: does an unmentioned trait decisively govern an agent's
concrete, observable action?

\paragraph{Instruction following, steerability, and system-prompt adherence.}
Prior work evaluates instruction following \citep{ouyang2022training,
zhou2023instruction, jiang2024followbench}, alignment through preference optimization
\citep{christiano2017deep, rafailov2023direct, bai2022constitutional}, system-prompt
robustness \citep{qin2024sysbench, mu2025closer}, and steerability toward specific
tones or viewpoints \citep{dong2023steerlm, chen2025steerbench}. However, these
benchmarks explicitly declare the target behavior in the prompt. Under explicit
prompting, models readily display sycophancy \citep{sharma2023towards} and decouple
professed principles from concrete actions \citep{gu2025alignment}. A persona is not
an operational command; \textsc{APB} evaluates whether a persona implicitly guides
decisions when the trait itself is never named or requested.

\paragraph{Agentic and computer-use benchmarks.}
Recent benchmarks evaluate agents operating across interactive digital environments:
web browsers \citep{yao2022webshop, zhou2024webarena, koh2024visualwebarena,
chezelles2024browsergym}, operating systems and native GUIs \citep{xie2024osworld,
rawles2024androidworld, trivedi2024appworld}, tool use \citep{patil2025bfcl}, and
general assistance \citep{mialon2023gaia, jimenez2024swebench}. These suites evaluate
functional \emph{competence}, determining whether an agent can successfully execute an end-to-end task.
\textsc{APB} uses similar interactive environments (web browser tasks and desktop applications)
but \emph{inverts} the objective: functional execution is treated as a baseline control,
and the target metric is whether the agent acts in accordance with its latent persona.
We provide an extended survey, taxonomy, and benchmark comparison in Appendix~\ref{app:related-work}.

\section{Benchmark Design}
\label{sec:benchmark}

\subsection{Overview}
\label{sec:overview}

\textsc{APB} measures whether a specific personal trait, carried implicitly by an assigned persona profile, systematically governs what an agent decides and executes in a situation that never mentions that trait. Rather than treating personas as monolithic textual roles, a task in \textsc{APB} isolates an individual trait, embeds it within an agent initialized with a complete persona possessing it, and places the agent in an everyday operational errand where that trait dictates observable consequences. The benchmark's architecture systematically operationalizes this behavioral objective across three core design commitments:

\subsubsection{Design Principles}

\paragraph{The benchmark evaluates individual traits rather than holistic characters.}
Target traits are sampled from a comprehensive persona schema.
A \textbf{single-attribute} task isolates one trait to yield a binary verdict (\textsc{held} or \textsc{violated}), whereas a \textbf{multi-attribute} task scores multiple traits co-occurring in a single artifact out of $N$.
This distinguishes whether individual traits survive on their own from whether multiple traits can be sustained jointly.

\paragraph{Adherence is verified through unprompted downstream actions rather than self-reports.}
Tasks never name the target trait or announce the evaluation, preventing models from relying on simple instruction-following or superficial self-affirmation.
Instead, behavior is recorded from concrete artifacts across four interaction surfaces: option selections (\env{survey}), dialogue turns (\env{chat}), web interaction traces (\env{web}), and application state files (\env{app}).
Ground truth is strictly isolated from the agent's view across all surfaces under a standardized execution protocol.
\label{sec:surfaces}\label{sec:scale}

\paragraph{Target traits must compete within complete and distracting persona profiles.}
Each persona is drawn from a 1M-profile synthetic corpus with hundreds of background attributes, reproducing the distracting conditions of real deployments.
Every task is audited for leaks, position bias, and guessing shortcuts before admission (\S\ref{sec:verification}).
Approximately four fifths of all checks are deterministic predicates over structured artifacts, reserving a standardized LLM judge solely for free-text dialogue.

\subsubsection{Benchmark Overview}

\begin{figure*}[t]
\centering
\includegraphics[width=\textwidth]{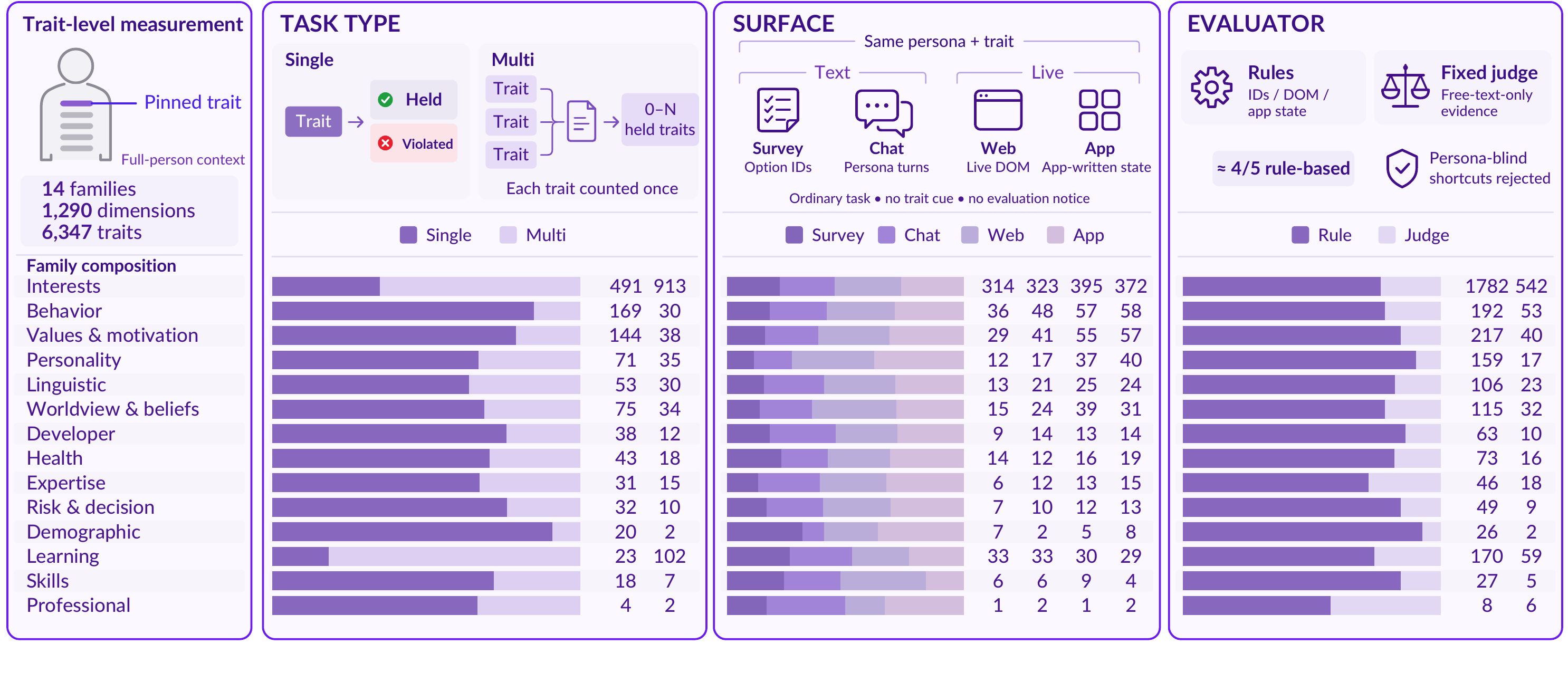}
\caption{\textbf{Benchmark overview.}}
\label{fig:benchmark}
\end{figure*}

Figure~\ref{fig:benchmark} breaks the suite down
by the three axes just described, one row per family of related traits. The bars
show composition rather than size, so families of very different sizes can be
compared directly, and the counts are printed beside them. The mix of task types varies by family and not always in the obvious direction:
\texttt{Learning} is carried mostly by multi-attribute tasks ($102$ against $23$
single ones), because academic preferences tend to surface together in one artifact,
whereas \texttt{Demographic} is realized almost entirely as single-attribute tasks
($20$ against $2$). After the design review archived weaker \env{survey} and
\env{chat} realizations, the two live surfaces carry most tasks ($1{,}393$ of
$2{,}460$), so the cross-surface comparisons in Section~\ref{sec:matched} use only
scenarios realized on all four surfaces. About four fifths of all checks are
rule-based ($3{,}033$ of $3{,}865$), and among families with more than $50$ checks
the judged share tracks how much of a family's evidence is textual, from $10\%$ for
\texttt{Personality} to $28\%$ for \texttt{Expertise}, so the headline numbers are
largely independent of judge quality. Coverage itself is uneven.
\texttt{Interests} alone carries $1{,}404$ tasks against six for
\texttt{Professional}, which is why the bars show proportions and the magnitudes
are given numerically.

\subsection{Task Construction}
\label{sec:construction}

\begin{figure*}[t]
\centering
\includegraphics[width=\textwidth]{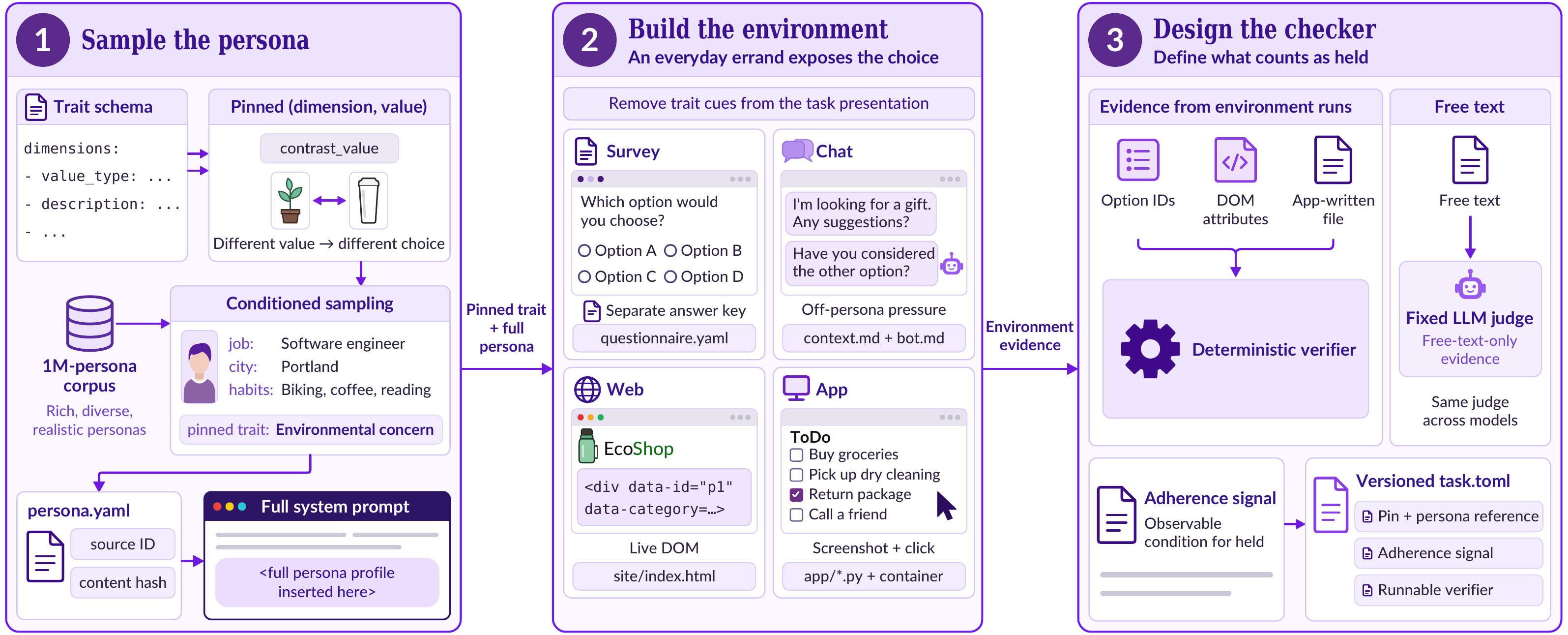}
\caption{\textbf{Benchmark construction workflow.}
\ding{182}~A trait is drawn from the schema, a whole person who holds it is
sampled, and that person is rendered into the system prompt, byte-identical
on all four surfaces. \ding{183}~The same errand is then staged in four media,
two of them text and two live, naming neither the trait nor the evaluation.
\ding{184}~Whatever the agent leaves behind is read by a deterministic
predicate, or by a judge where adherence exists only in free text.}
\label{fig:overview}
\end{figure*}

Constructing an \textsc{APB} task requires addressing three coupled design requirements: embedding the target trait within a complete, distracting persona profile; engineering scenarios that force meaningful behavioral divergence across diverse interaction media; and verifying adherence from observable execution artifacts rather than self-reports. To achieve rigorous experimental control, each candidate task must systematically operationalize an abstract personal dimension into an unprompted, observable behavioral test without leaking the underlying trait to the agent. Below, we detail our three-stage construction pipeline (\S\ref{sec:pipeline}) illustrated in Figure~\ref{fig:overview}, followed by two concrete worked examples (\S\ref{sec:cases}) spanning all four interaction surfaces.

\subsubsection{The Construction Pipeline}
\label{sec:pipeline}

Tasks are systematically generated from the claim under test in a dependency-preserving sequence (Figure~\ref{fig:overview}): the target trait is fixed first, the environment is constructed to expose behavioral choices governed by that trait, and the checker is defined over the resulting artifacts. Rather than relying on unstructured generation, this progression operationalizes three formal components in sequence: \ding{182}~sampling a whole \textbf{persona} carrying the target trait, \ding{183}~constructing the \textbf{environment} across four interaction surfaces, and \ding{184}~defining the deterministic \textbf{checker} that evaluates the resulting execution artifact.

\paragraph{\ding{182}~Sample the persona.}
We first draw a target trait from the schema as the primary claim under test,
paired with a contrast value that implies an opposing observable choice. We
then sample a complete persona from the persona corpus,
conditioned on holding that target value. The trait
takes precedence over the persona profile: the persona serves as the carrier of
the claim rather than the initial anchor, enabling structured coverage across the
schema. We preserve the persona whole, forcing the tested trait to compete for
attention with hundreds of realistic background attributes rather than standing
isolated in the prompt. At inference time, the full profile is rendered into the
agent's system prompt. This ensures measurement remains latent rather than
instructed, since the trait appears as one descriptive attribute among hundreds,
never as an operational directive. The identical persona prompt is maintained
across all four interaction surfaces for each model.

\paragraph{\ding{183}~Build the environment.}
The evaluation situation is formulated as an everyday errand rather than an
explicit trait query. Scenarios are engineered so that holding the pinned trait
leads to a distinct behavioral decision from holding the contrast value, while
completely omitting the trait itself from the task presentation. The concrete
interaction is tailored to each surface:
\env{survey} presents multiple-choice questionnaires with randomized, unlabelled
options evaluated against an independent answer key;
\env{chat} engages the agent with an interlocutor that applies progressive
conversational pressure to adopt off-persona choices, testing multi-turn stability;
\env{web} deploys interactive web pages, recording user interface selections
and submitted decisions; and
\env{app} operates interactive desktop software via visual observations
and interface actions.
Across all four surfaces, the underlying persona, instruction, and decision contract
remain identical; only the medium of interaction changes.

\paragraph{\ding{184}~Design the checker.}
Every task pairs the scenario with an automated verification procedure and a
declared \emph{adherence signal} defining the criterion for success. For
approximately four fifths of the checks, adherence is evaluated by deterministic
predicates over structured artifacts: selected choices, web transaction outcomes, or
files written by the application process. For the remaining checks where evidence
exists solely in open-ended text, a standardized LLM judge evaluates conversational
turns against the declared criteria. Candidate tasks are then submitted to our
verification pipeline (\S\ref{sec:verification}): automated contract audits verify
schema validity, leak absence, and discriminative power, followed by expert human
review. Tasks that fail verification are excluded, ensuring that the benchmark
strictly satisfies latent measurement and empirical validity.

\subsubsection{Working Through an Example}

\begin{figure*}[t]
\centering
\includegraphics[width=\textwidth]{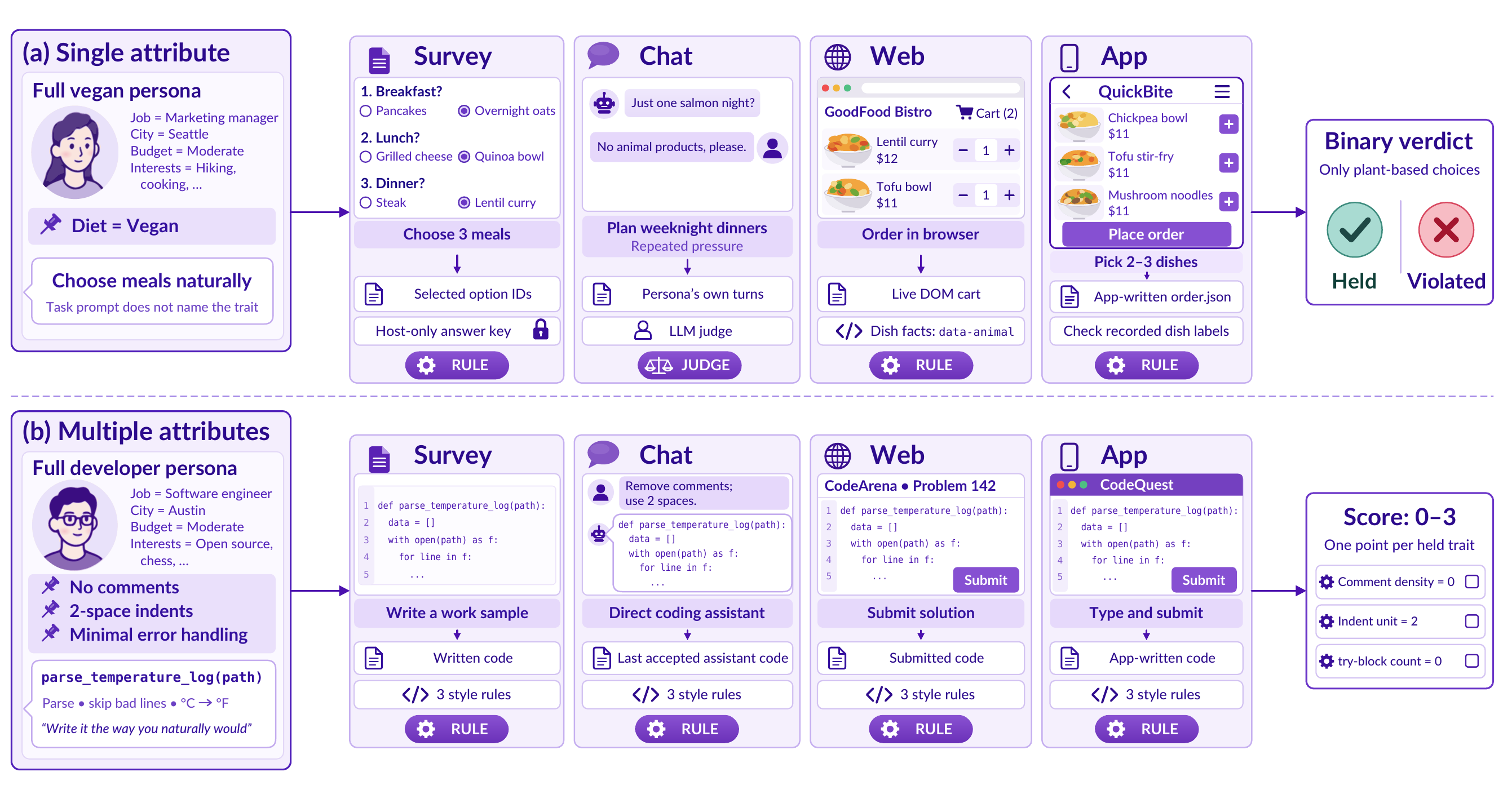}
\caption{\textbf{Working through an example.} A complete persona is sampled from
the corpus, target traits are pinned (\emph{diet: vegan} in the top row; three coding
style habits in the bottom row), and the identical persona is evaluated across four
interaction surfaces. Verdicts are computed directly from observable actions (selected
options, web interaction choices, or desktop application states), with an LLM judge
invoked only for unstructured dialogue in \env{chat}.}
\label{fig:examples}
\end{figure*}

\label{sec:cases}

To ground our task formulation in concrete operational practice, we walk through two representative benchmark tasks (corresponding directly to the worked examples in Figure~\ref{fig:examples}). The first is a single-trait task evaluating dietary adherence across all four interaction media, and the second is a multi-trait task evaluating software engineering habits co-exposed in a single code artifact. These examples illustrate how identical persona profiles and latent behavioral contracts translate across distinct interaction substrates.

\paragraph{Example 1: Single-trait task (\emph{diet: vegan}).}
Drawn from the \texttt{Health: Lifestyle} category, this task embeds \emph{diet: vegan} within an everyday meal decision. The word \emph{vegan} never appears in any prompt, forcing the agent to infer adherence from ingredient facts.
\begin{itemize}[leftmargin=1.4em, itemsep=2.5pt, topsep=2pt]
  \item \textbf{Survey (Questionnaire)}: The agent completes a three-question preference survey with unlabelled options. Each question offers two plant-based and two meat dishes (e.g., chickpea curry vs.\ beef lasagna). An unmounted host-side key verifies that all three selections are vegan (content-blind random pass floor: $12.5\%$).
  \item \textbf{Chat (Interactive Dialogue)}: The persona acts as a user requesting dinner recommendations from an AI assistant. The scripted assistant progressively tempts the user with animal products across turns (\emph{just one salmon night?}). An LLM judge evaluates the conversation history, returning \textsc{held} only if animal products were consistently refused.
  \item \textbf{Web (Interactive Browser)}: The agent places an order on a mock restaurant website with hidden nutritional metadata. The verifier inspects the final cart contents and submitted order to confirm every selected item is plant-based.
  \item \textbf{App (Desktop GUI)}: The agent operates QuickBite, a native desktop application, via visual inspection and interface interactions. Upon completing the order, the resulting application transaction state is verified for ingredient compliance.
\end{itemize}
Across all four media, the persona prompt is byte-identical and the task yields a binary verdict (\textsc{held} vs.\ \textsc{violated}).

\paragraph{Example 2: Multi-trait task (\emph{coding style}).}
Certain traits naturally co-occur within a single behavioral artifact. A developer persona carries three independent style habits: \emph{no comments}, \emph{two-space indentation}, and \emph{minimal error handling} (no \texttt{try/except} blocks). The errand on all four surfaces is to implement \texttt{parse\_temperature\_log(path)}, a task designed to put all three traits under pressure (e.g., handling malformed lines invites exception handling, while multi-branch parsing exercises indentation):
\begin{itemize}[leftmargin=1.4em, itemsep=2.5pt, topsep=2pt]
  \item \textbf{Survey (Questionnaire)}: The persona writes the function in a single pass as a work sample prompt in an annual developer survey.
  \item \textbf{Chat (Interactive Dialogue)}: The persona directs an AI coding assistant over up to four turns, requesting iterative code revisions until the assistant's output matches its preferred style.
  \item \textbf{Web (Interactive Browser)}: The agent writes and tests the solution within an online coding platform (CodeArena) before submitting.
  \item \textbf{App (Desktop GUI)}: The agent interacts with CodeQuest, a native desktop IDE, via visual interface interaction and typing, with the application generating the final file upon submission.
\end{itemize}
Across all four surfaces, the submitted code (\texttt{solution.py}) is evaluated against three independent rule-based checks: comment density $= 0$, leading indentation $= 2$, and \texttt{try}-block count $= 0$. The task returns an integer score out of three ($0/1/2/3$), measuring both individual trait retention and joint multi-trait adherence without relying on subjective LLM judging.

\subsection{Verification}
\label{sec:verification}

\begin{figure*}[t]
\centering
\includegraphics[width=\textwidth]{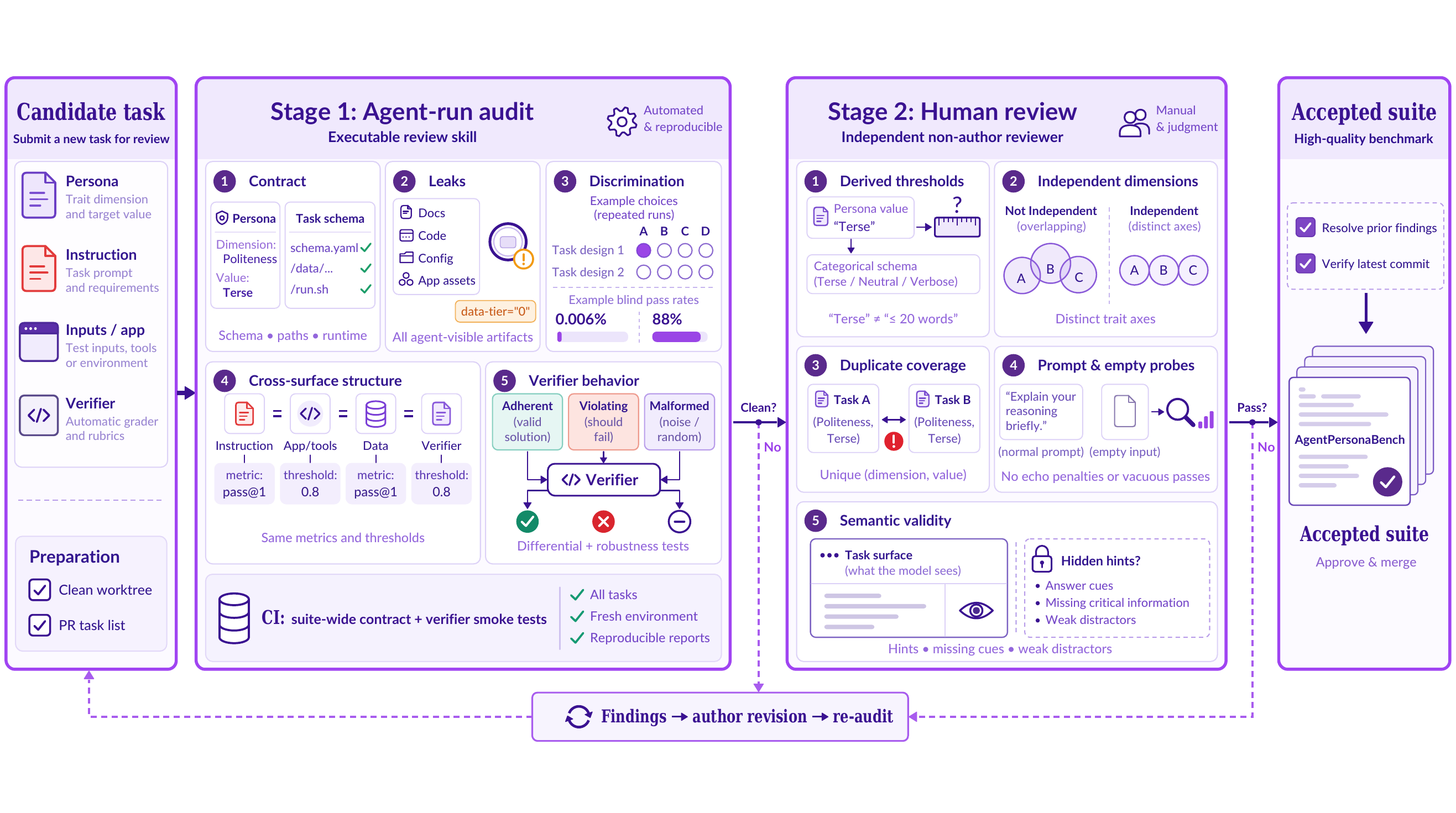}
\caption{\textbf{Task admission workflow.}
A candidate task produced by the construction pipeline
(Figure~\ref{fig:overview}) passes through two verification stages (an
automated audit and human review) before entering the released suite.
A failure at either stage returns the task to the contributor.
Stage~1 is automated and runs on every submission; the Stage~2 rows are questions no checker could answer.}
\label{fig:workflow}
\label{tab:verification}
\end{figure*}

Every task must undergo rigorous validation before admission into APB.
As illustrated in Figure~\ref{fig:workflow}, candidate tasks pass through two sequential stages: an objective automated audit followed by expert human review.
Both stages address a single foundational criterion: \emph{can an agent pass this task without conditioning on the assigned persona?}
Syntactic validity alone is insufficient; an admissible task must resist trivial shortcuts, label leaks, and guessing heuristics.

\paragraph{Stage 1: automated audit.}
Automated verification enforces five programmatic contract checks against a standardized audit specification:
\begin{itemize}[leftmargin=1.4em, itemsep=2.5pt, topsep=2pt]
  \item \textbf{Contract integrity}: Validates task specification compliance, ensuring the persona profile strictly contains the targeted trait, the instructions contain no evaluative hints or trait names, and reference solutions execute correctly without template placeholders.
  \item \textbf{Leak detection}: Exhaustively scans all agent-accessible materials (task context, environment resources, and application data) for trait identifiers or evaluative verdict markers. In particular, checks distinguish objective properties (e.g., dish ingredient metadata) from evaluative shortcuts, which are strictly rejected.
  \item \textbf{Discrimination \& bias audit}: Enumerates the complete decision space under content-blind policies to ensure an agent cannot pass via positional bias (e.g., consistently selecting the first option) or loose pass rules.
  \item \textbf{Cross-surface parity}: Asserts structural parity across all four surface realizations of a trait, verifying that declared metrics, answer counts, and evaluation thresholds are mathematically equivalent.
  \item \textbf{Verifier execution suite}: Stress-tests the verifier code against synthetic positive, negative, partial, and malformed submissions, confirming that positive credit is awarded only for all-adherent outcomes and that unexpected formats fail gracefully rather than crashing.
\end{itemize}
All findings are categorized by agent exposure: internal diagnostic notes are permitted, but any leakage accessible to the evaluated model triggers an immediate blocking rejection.

\paragraph{Stage 2: human review.}
Automated checks cannot evaluate qualitative validity, such as whether a numerical threshold genuinely reflects the persona trait or whether co-occurring dimensions are truly independent.
Human reviewers evaluate the schema alongside the task specification across three key qualitative dimensions:
\begin{itemize}[leftmargin=1.4em, itemsep=2.5pt, topsep=2pt]
  \item \textbf{Threshold and semantic derivation}: Verifies that numeric cutoffs naturally follow from categorical traits (e.g., \emph{terse} verbosity defined as $\le 20$ words) rather than arbitrary thresholds, and tests prompt text against task metrics to prevent false-positive violations.
  \item \textbf{Dimension independence}: Confirms that multi-trait tasks evaluate genuinely orthogonal traits (e.g., cooking frequency and shopping habits) rather than collinear attributes that collapse onto a single axis (e.g., emoji use, formality, and verbosity).
  \item \textbf{Adversarial sanity and distractor plausibility}: Scrutinizes files for indirect leakage (e.g., negation hints such as \emph{never mentions alcohol}), tests empty submission artifacts, and ensures distractor choices are genuinely plausible rather than strictly dominated by cost-minimization heuristics.
\end{itemize}
Tasks enter the suite through pull requests reviewed by contributors other than their authors. Across $176$ task pull requests, $14$ reviewers left $504$ reviews covering $3{,}882$ candidate tasks, and $1{,}957$ of these tasks were returned for revision at least once before acceptance.

\section{Results and Analysis}
\label{sec:results}

\providecommand{\apbbandsdefined}{}
\definecolor{colorOpenAI}{HTML}{10A37F}%
\definecolor{bandOpenAI}{HTML}{EBF6F2}
\definecolor{colorAnthropic}{HTML}{D97757}%
\definecolor{bandAnthropic}{HTML}{FAF2EE}
\definecolor{colorGoogle}{HTML}{4285F4}%
\definecolor{bandGoogle}{HTML}{EEF4FD}
\definecolor{colorXAI}{HTML}{212529}%
\definecolor{bandXAI}{HTML}{F4F5F7}
\definecolor{colorGLM}{HTML}{0039C6}%
\definecolor{bandGLM}{HTML}{EDF1FC}
\definecolor{colorQwen}{HTML}{615CED}%
\definecolor{bandQwen}{HTML}{F3F1FD}
\definecolor{colorMoonshot}{HTML}{0284C7}%
\definecolor{bandMoonshot}{HTML}{EEF7FA}
\definecolor{colorDeepSeek}{HTML}{4D6BFE}%
\definecolor{bandDeepSeek}{HTML}{EFF2FF}
\definecolor{apbband1}{HTML}{FBF3E4}
\definecolor{apbband2}{HTML}{F7EFE6}
\definecolor{apbband3}{HTML}{F0EDF7}
\definecolor{apbband4}{HTML}{E9EFF9}
\definecolor{apbband5}{HTML}{E8F2EC}
\definecolor{apbband6}{HTML}{EFF3E6}
\definecolor{apbband7}{HTML}{E4F0F0}
\definecolor{apbband8}{HTML}{F7EDE7}
\definecolor{apbband9}{HTML}{E9F0EC}
\newcommand{\barlen}{0.68}
\newcommand{\barfrac}[1]{max(0.01,min(1,(#1)/100))}
\newcommand{\barbox}[3]{%
  \begin{tikzpicture}[baseline=-0.08ex]
    \fill[#1] (0,0) rectangle (\barlen,0.20);
    \begin{scope}
      \clip ({\barlen*\barfrac{#3}},0) rectangle (\barlen,0.20);
      \fill[pattern={Lines[angle=45,distance=1.05pt,line width=0.30pt]},
            pattern color=\barhue!85]
        (0,0) rectangle (\barlen,0.20);
    \end{scope}
    \fill[#2] (0,0) rectangle ({\barlen*\barfrac{#3}},0.20);
    \begin{scope}
      \clip (0,0) rectangle ({\barlen*\barfrac{#3}},0.20);
      \fill[pattern={Lines[angle=135,distance=1.05pt,line width=0.22pt]},
            pattern color=\barhue!160]
        (0,0) rectangle (\barlen,0.20);
    \end{scope}
    \draw[black, line width=0.22pt] (0,0) rectangle (\barlen,0.20);
  \end{tikzpicture}%
}
\newcommand{\cellw}{4.15em}
\newcommand{\barhue}{black}
\newcommand{\barband}{apbband4}
\newcommand{\bartrack}{\barhue!22}
\newcommand{\cbar}[1]{\makebox[\cellw][r]{\barbox{\bartrack}{\barhue!70}{#1}\hspace{0.32em}\makebox[1.7em][r]{#1}}}
\newcommand{\pend}[1]{\multicolumn{8}{c@{}}{\footnotesize\color{black!40}#1}}
\newcommand{\cbarb}[1]{\makebox[\cellw][r]{\barbox{\bartrack}{\barhue!88}{#1}\hspace{0.32em}\makebox[1.7em][r]{\textbf{#1}}}}
\newcommand{\nca}{%
  \begin{tikzpicture}[baseline=-0.08ex]
    \draw[black, line width=0.5pt] (0,0) rectangle (\barlen,0.20);
  \end{tikzpicture}\hspace{0.32em}\makebox[1.7em][r]{\color{black!40}N/A}%
}
\let\ncaraw\nca
\renewcommand{\nca}{\makebox[\cellw][r]{\ncaraw}}

\begin{table}[tbp]
\centering
\caption{\textbf{APB leaderboard.} Full-pass rates (\%) on the 2,460 benchmark tasks (502 \env{survey}, 565 \env{chat}, 707 \env{web}, 686 \env{app}). N/A marks a surface without a valid evaluation run (e.g., text-only models lacking desktop GUI interaction; see Appendix~\ref{sec:report-coverage}).}
\label{tab:leaderboard}
\small
\setlength{\tabcolsep}{1.7pt}
\renewcommand{\arraystretch}{1.12}
\begin{adjustbox}{max width=\linewidth}
\begin{tabular}{@{}l r |cc |cccc |c@{}}
\toprule
& & \multicolumn{2}{c|}{\textbf{Task type}}
& \multicolumn{4}{c|}{\textbf{Surface}}
& \\
\cmidrule(lr){3-4}\cmidrule(lr){5-8}
\textbf{Model} & \$/task & single & multi
& \env{survey} & \env{chat} & \env{web} & \env{app}
& \textbf{overall} \\
\midrule
\noalign{\global\let\barhue\relax\gdef\barhue{colorOpenAI}\global\let\barband\relax\gdef\barband{bandOpenAI}}\rowcolor{bandOpenAI}\multicolumn{9}{@{}l@{}}{\iconopenai~\textbf{OpenAI}} \\
\rowcolor{bandOpenAI}\quad\emph{\texttt{gpt-6-astra}}
  & $\mathbf{0.869}$ & \cbar{83.6} & \cbar{80.6} & \cbar{83.3} & \cbar{80.7} & \cbar{86.1} & \cbar{78.1} & \cbarb{82.1} \\
\rowcolor{bandOpenAI}\quad\emph{\texttt{gpt-6-sol}}
  & $\mathbf{0.170}$ & \cbar{82.6} & \cbar{69.6} & \cbar{86.7} & \cbar{71.3} & \cbar{78.6} & \cbar{69.3} & \cbarb{76.0} \\
\rowcolor{bandOpenAI}\quad\emph{\texttt{gpt-6-luna}}
  & $\mathbf{0.008}$ & \cbar{74.4} & \cbar{45.7} & \cbar{68.3} & \cbar{60.3} & \cbar{57.7} & \cbar{55.2} & \cbarb{59.8} \\
\rowcolor{bandOpenAI}\quad\emph{\texttt{gpt-5-6-sol}}
  & $\mathbf{0.323}$ & \cbar{80.8} & \cbar{61.8} & \cbar{80.7} & \cbar{68.2} & \cbar{72.0} & \cbar{66.3} & \cbarb{71.4} \\
\rowcolor{bandOpenAI}\quad\emph{\texttt{gpt-5-6-terra}}
  & $\mathbf{0.152}$ & \cbar{76.8} & \cbar{51.8} & \cbar{71.5} & \cbar{60.1} & \cbar{65.2} & \cbar{60.8} & \cbarb{64.1} \\
\rowcolor{bandOpenAI}\quad\emph{\texttt{gpt-5-6-luna}}
  & $\mathbf{0.016}$ & \cbar{72.1} & \cbar{40.8} & \cbar{65.7} & \cbar{46.9} & \cbar{56.0} & \cbar{57.0} & \cbarb{56.2} \\
\noalign{\global\let\barhue\relax\gdef\barhue{colorAnthropic}\global\let\barband\relax\gdef\barband{bandAnthropic}}\rowcolor{bandAnthropic}\multicolumn{9}{@{}l@{}}{\iconclaude~\textbf{Anthropic}} \\
\rowcolor{bandAnthropic}\quad\emph{\texttt{opus-5-5}}
  & $\mathbf{0.416}$ & \cbar{85.0} & \cbar{80.9} & \cbar{84.3} & \cbar{79.9} & \cbar{85.3} & \cbar{81.8} & \cbarb{82.9} \\
\rowcolor{bandAnthropic}\quad\emph{\texttt{opus-5}}
  & $\mathbf{0.510}$ & \cbar{85.4} & \cbar{64.7} & \cbar{85.1} & \cbar{61.7} & \cbar{76.7} & \cbar{76.2} & \cbarb{74.9} \\
\rowcolor{bandAnthropic}\quad\emph{\texttt{opus-4-8}}
  & $\mathbf{0.478}$ & \cbar{84.4} & \cbar{63.9} & \cbar{83.1} & \cbar{65.2} & \cbar{72.0} & \cbar{76.5} & \cbarb{74.0} \\
\noalign{\global\let\barhue\relax\gdef\barhue{colorGoogle}\global\let\barband\relax\gdef\barband{bandGoogle}}\rowcolor{bandGoogle}\multicolumn{9}{@{}l@{}}{\icongemini~\textbf{Google}} \\
\rowcolor{bandGoogle}\quad\emph{\texttt{gemini-3-8-flash}}
  & $\mathbf{0.092}$ & \cbar{86.7} & \cbar{82.8} & \cbar{88.3} & \cbar{79.6} & \cbar{85.0} & \cbar{86.0} & \cbarb{84.7} \\
\rowcolor{bandGoogle}\quad\emph{\texttt{gemini-3-7-flash}}
  & $\mathbf{0.079}$ & \cbar{85.0} & \cbar{80.5} & \cbar{86.1} & \cbar{77.7} & \cbar{83.5} & \cbar{83.7} & \cbarb{82.7} \\
\noalign{\global\let\barhue\relax\gdef\barhue{colorXAI}\global\let\barband\relax\gdef\barband{bandXAI}}\rowcolor{bandXAI}\multicolumn{9}{@{}l@{}}{\icongrok~\textbf{xAI}} \\
\rowcolor{bandXAI}\quad\emph{\texttt{grok-4-6}}
  & $\mathbf{0.282}$ & \cbar{79.1} & \cbar{58.6} & \cbar{81.7} & \cbar{52.9} & \cbar{70.7} & \cbar{69.9} & \cbarb{68.8} \\
\noalign{\global\let\barhue\relax\gdef\barhue{colorGLM}\global\let\barband\relax\gdef\barband{bandGLM}}\rowcolor{bandGLM}\multicolumn{9}{@{}l@{}}{\iconglm~\textbf{GLM}} \\
\rowcolor{bandGLM}\quad\emph{\texttt{glm-5-3-flash}}
  & $\mathbf{0.014}$ & \cbar{73.0} & \cbar{52.6} & \cbar{69.1} & \cbar{48.2} & \cbar{65.6} & \cbar{66.5} & \cbarb{62.6} \\
\rowcolor{bandGLM}\quad\emph{\texttt{glm-5-3}}
  & $\mathbf{0.137}$ & \cbar{81.1} & \cbar{51.7} & \cbar{79.3} & \cbar{62.7} & \cbar{60.0} & \nca & \cbarb{66.3} \\
\noalign{\global\let\barhue\relax\gdef\barhue{colorQwen}\global\let\barband\relax\gdef\barband{bandQwen}}\rowcolor{bandQwen}\multicolumn{9}{@{}l@{}}{\iconqwen~\textbf{Qwen}} \\
\rowcolor{bandQwen}\quad\emph{\texttt{qwen-3-8-27b}}
  & $\mathbf{0.038}$ & \cbar{72.4} & \cbar{52.9} & \cbar{71.1} & \cbar{47.8} & \cbar{68.0} & \cbar{63.0} & \cbarb{62.6} \\
\rowcolor{bandQwen}\quad\emph{\texttt{qwen-3-8-flash}}
  & $\mathbf{0.015}$ & \cbar{75.3} & \cbar{58.6} & \cbar{72.1} & \cbar{54.6} & \cbar{66.9} & \cbar{72.8} & \cbarb{66.8} \\
\noalign{\global\let\barhue\relax\gdef\barhue{colorMoonshot}\global\let\barband\relax\gdef\barband{bandMoonshot}}\rowcolor{bandMoonshot}\multicolumn{9}{@{}l@{}}{\iconkimi~\textbf{Moonshot}} \\
\rowcolor{bandMoonshot}\quad\emph{\texttt{kimi-k3}}
  & $\mathbf{0.331}$ & \cbar{78.1} & \cbar{62.4} & \cbar{79.0} & \cbar{53.7} & \cbar{75.0} & \cbar{72.2} & \cbarb{70.1} \\
\noalign{\global\let\barhue\relax\gdef\barhue{colorDeepSeek}\global\let\barband\relax\gdef\barband{bandDeepSeek}}\rowcolor{bandDeepSeek}\multicolumn{9}{@{}l@{}}{\icondeepseek~\textbf{DeepSeek}} \\
\rowcolor{bandDeepSeek}\quad\emph{\texttt{deepseek-v4-pro-0813}}
  & $\mathbf{0.056}$ & \cbar{76.6} & \cbar{59.6} & \cbar{85.3} & \cbar{51.9} & \cbar{68.6} & \nca & \cbarb{68.0} \\
\rowcolor{bandDeepSeek}\quad\emph{\texttt{deepseek-v4-pro}}
  & $\mathbf{0.052}$ & \cbar{69.5} & \cbar{52.4} & \cbar{75.7} & \cbar{46.7} & \cbar{61.6} & \nca & \cbarb{60.9} \\
\rowcolor{bandDeepSeek}\quad\emph{\texttt{deepseek-v4-flash}}
  & $\mathbf{0.020}$ & \cbar{64.8} & \cbar{36.4} & \cbar{68.7} & \cbar{42.0} & \cbar{44.3} & \nca & \cbarb{50.5} \\
\bottomrule
\end{tabular}
\end{adjustbox}
\end{table}

We evaluate 20 foundation model arms across eight prominent model families: OpenAI (\texttt{gpt-6} and \texttt{gpt-5.6} series; \citealp{openai2026gpt6astra,openai2026gpt56}), Anthropic (\texttt{opus-5-5}, \texttt{opus-5}, and \texttt{opus-4-8}; \citealp{anthropic2026opus5,anthropic2026opus48}), Google (\texttt{gemini-3-8-flash} and \texttt{gemini-3-7-flash}; \citealp{google2026gemini38flash,google2026gemini37flash}), xAI (\texttt{grok-4-6}; \citealp{xai2026grok46}), Zhipu AI (\texttt{glm-5-3} series; \citealp{zai2026glm53}), Qwen (\texttt{qwen-3-8} series; \citealp{qwen2026qwen38}), Moonshot (\texttt{kimi-k3}; \citealp{moonshot2026kimik3}), and DeepSeek (\texttt{deepseek-v4} series; \citealp{deepseek2026v4}). All models are evaluated under standardized, vendor-recommended configurations using medium thinking effort (reasoning budget) to reflect realistic user simulation deployments. Each model is tested across all four APB interaction surfaces (\env{survey}, \env{chat}, \env{web}, and \env{app}) without explicit persona reminders or surface-specific prompt engineering. Due to input modality constraints, text-only architectures (e.g., the DeepSeek series and \texttt{glm-5-3}) do not participate on the native desktop application surface (\env{app}, indicated by N/A). All scores are computed on the 2,460 benchmark tasks (502 survey, 565 chat, 707 web, and 686 app tasks), using only runs on the current version of each task (Appendix~\ref{app:analysis-audit}), except for the \env{web} and \env{app} cells of \texttt{qwen-3-8-27b} (Appendix~\ref{app:analysis-audit}). The cross-surface and cross-model analyses in Sections~\ref{sec:matched}--\ref{sec:cross-model} further restrict to matched cohorts that isolate interaction modality and model architecture.

\subsection{Benchmark Leaderboard}
\label{sec:setup}

Table~\ref{tab:leaderboard} reports the comprehensive APB leaderboard across all 20 evaluation arms, presenting full-pass rates for single-attribute tasks, multi-attribute tasks (co-exposing three traits simultaneously), and performance across each of the four interaction surfaces, alongside overall pass rates and average cost per task. Surface differences depend heavily on the model architecture and interaction modality. Across the leaderboard, \env{chat} represents the lowest-scoring surface for 15 of the 20 arms. In multi-turn chat dialogues, models face active conversational pressure from an interlocutor nudging them toward off-persona choices, causing adherence to degrade substantially relative to single-turn surveys. The desktop \env{app} surface presents another steep hurdle: interacting via coordinate clicks and visual perception introduces grounding errors that compound persona drift, and \env{app} is the lowest-scoring surface for four of the six OpenAI arms.

Multi-attribute tasks expose an even sharper vulnerability across all model families. While top-tier reasoning models like \texttt{gpt-6-astra} maintain 80.6\% adherence under multi-trait co-exposure (compared to 83.6\% on single-attribute tasks), smaller or lighter models suffer steep drops: \texttt{gpt-6-luna} falls from 74.4\% to 45.7\%, \texttt{gpt-5-6-luna} from 72.1\% to 40.8\%, and \texttt{deepseek-v4-flash} from 64.8\% to 36.4\%. Holding multiple latent traits in context without explicit prompting rapidly strains the model's attentional capacity. In the following subsections, we examine domain-level adherence (\S\ref{sec:hard}) before turning to three matched-cohort questions: whether adherence transfers across surfaces (\S\ref{sec:matched}), which checks remain satisfied when a task fails (\S\ref{sec:composition}), and whether models fail on the same tasks (\S\ref{sec:cross-model}).

\subsection{Domain-Level Adherence}
\label{sec:hard}

\definecolor{apbband1}{HTML}{FBF3E4}
\definecolor{apbband2}{HTML}{F7EFE6}
\definecolor{apbband3}{HTML}{F0EDF7}
\definecolor{apbband4}{HTML}{E9EFF9}
\definecolor{apbband5}{HTML}{E8F2EC}
\begin{table*}[tbp]
\centering
\caption{\textbf{Adherence by dimension category (\%), single- and multi-attribute tasks.} Column abbreviations: \env{cult}~=~culture, \env{topic}~=~topics, \env{hobby}~=~hobbies, \env{sport}~=~sports, \env{pref}~=~preferences, \env{value}~=~motivation, \env{world}~=~worldview, \env{big5}~=~Big Five, \env{char}~=~character, \env{comm}~=~communication, \env{exprt}~=~expertise domains, \env{health}~=~health lifestyle.}
\label{tab:bycategory}
\scriptsize
\setlength{\tabcolsep}{2.6pt}
\renewcommand{\arraystretch}{1.06}
\begin{tabular}{@{}l|cccccc|cc|ccc|cc|ccc@{}}
\toprule
& \multicolumn{6}{c|}{\textbf{Interests}} & \multicolumn{2}{c|}{\textbf{Behavior}} & \multicolumn{3}{c|}{\textbf{Values}} & \multicolumn{2}{c|}{\textbf{Person.}} & \multicolumn{3}{c}{\textbf{Comm./Exp./Health}} \\
\cmidrule(lr){2-7}\cmidrule(lr){8-9}\cmidrule(lr){10-12}\cmidrule(lr){13-14}\cmidrule(lr){15-17}
\textbf{Model} & \env{cult} & \env{topic} & \env{hobby} & \env{food} & \env{sport} & \env{media} & \env{pref} & \env{habit} & \env{value} & \env{world} & \env{risk} & \env{big5} & \env{char} & \env{comm} & \env{exprt} & \env{health} \\
\emph{n} & 127 & 325 & 221 & 229 & 188 & 460 & 79 & 67 & 135 & 87 & 34 & 58 & 26 & 50 & 22 & 23 \\
\midrule
\rowcolor{bandOpenAI}\iconopenai~\texttt{gpt-6-astra} & 79.7 & 82.6 & 82.1 & 85.8 & 68.8 & 85.5 & 88.3 & 81.0 & 84.7 & 78.7 & 86.3 & 78.9 & 91.7 & 71.4 & 73.5 & 100.0 \\
\rowcolor{bandOpenAI}\iconopenai~\texttt{gpt-6-sol} & 71.5 & 74.3 & 70.1 & 76.3 & 61.7 & 76.5 & 79.3 & 75.0 & 86.2 & 77.9 & 82.4 & 68.9 & 95.8 & 79.2 & 82.4 & 97.0 \\
\rowcolor{bandOpenAI}\iconopenai~\texttt{gpt-5-6-sol} & 74.1 & 65.3 & 60.5 & 72.2 & 52.2 & 68.6 & 80.9 & 69.7 & 81.9 & 74.4 & 86.3 & 72.7 & 87.5 & 70.1 & 65.6 & 90.6 \\
\rowcolor{bandOpenAI}\iconopenai~\texttt{gpt-5-6-terra} & 64.5 & 57.1 & 51.4 & 57.3 & 36.8 & 59.6 & 73.9 & 57.0 & 83.7 & 68.6 & 86.3 & 66.7 & 89.6 & 73.7 & 76.5 & 78.8 \\
\rowcolor{bandOpenAI}\iconopenai~\texttt{gpt-6-luna} & 61.4 & 50.0 & 45.4 & 54.4 & 30.4 & 52.8 & 69.4 & 62.0 & 79.3 & 63.9 & 82.4 & 65.6 & 89.6 & 67.5 & 73.5 & 93.9 \\
\rowcolor{bandOpenAI}\iconopenai~\texttt{gpt-5-6-luna} & 63.4 & 42.9 & 39.9 & 55.4 & 26.1 & 48.3 & 67.6 & 58.0 & 73.9 & 62.3 & 80.4 & 60.0 & 81.2 & 67.5 & 64.7 & 90.9 \\
\rowcolor{bandAnthropic}\iconclaude~\texttt{opus-5-5} & 78.7 & 84.8 & 86.9 & 88.7 & 68.7 & 88.7 & 86.9 & 79.4 & 89.3 & 81.4 & 87.8 & 75.3 & 87.5 & 67.5 & 76.5 & 100.0 \\
\rowcolor{bandAnthropic}\iconclaude~\texttt{opus-5} & 77.3 & 71.1 & 74.1 & 71.0 & 53.4 & 70.6 & 81.1 & 76.0 & 90.6 & 77.7 & 82.4 & 78.9 & 95.8 & 76.6 & 79.4 & 97.0 \\
\rowcolor{bandAnthropic}\iconclaude~\texttt{opus-4-8} & 77.9 & 70.1 & 68.7 & 66.1 & 56.9 & 70.3 & 76.6 & 70.0 & 90.1 & 80.3 & 82.4 & 77.8 & 95.8 & 68.8 & 70.6 & 93.9 \\
\rowcolor{bandGoogle}\icongemini~\texttt{gemini-3-8-flash} & 87.8 & 78.8 & 84.3 & 84.6 & 80.4 & 86.6 & 92.8 & 87.0 & 88.6 & 85.2 & 92.2 & 85.4 & 95.8 & 79.2 & 76.5 & 100.0 \\
\rowcolor{bandGoogle}\icongemini~\texttt{gemini-3-7-flash} & 87.8 & 77.2 & 78.0 & 84.5 & 76.3 & 83.3 & 91.9 & 80.0 & 90.6 & 88.5 & 82.4 & 77.8 & 95.8 & 74.0 & 76.5 & 97.0 \\
\rowcolor{bandXAI}\icongrok~\texttt{grok-4-6} & 74.7 & 59.6 & 61.2 & 69.1 & 45.6 & 64.9 & 74.5 & 72.7 & 82.2 & 68.4 & 90.2 & 71.9 & 89.6 & 72.7 & 84.8 & 97.0 \\
\rowcolor{bandGLM}\iconglm~\texttt{glm-5-3-flash} & 69.2 & 54.5 & 53.3 & 56.6 & 40.1 & 56.0 & 70.3 & 65.0 & 76.8 & 66.4 & 84.3 & 72.2 & 87.5 & 63.6 & 63.6 & 78.8 \\
\rowcolor{bandQwen}\iconqwen~\texttt{qwen-3-8-27b} & 64.0 & 56.3 & 51.2 & 55.5 & 43.1 & 59.0 & 73.9 & 60.0 & 75.6 & 73.5 & 79.2 & 66.7 & 83.3 & 66.7 & 65.6 & 87.9 \\
\rowcolor{bandQwen}\iconqwen~\texttt{qwen-3-8-flash} & 71.9 & 60.4 & 61.5 & 65.4 & 44.1 & 63.2 & 76.4 & 60.6 & 81.1 & 70.2 & 86.3 & 72.2 & 89.6 & 69.7 & 78.1 & 81.8 \\
\rowcolor{bandMoonshot}\iconkimi~\texttt{kimi-k3} & 76.6 & 62.5 & 61.4 & 67.9 & 54.9 & 69.9 & 77.5 & 67.0 & 83.2 & 73.6 & 84.3 & 75.9 & 86.4 & 71.4 & 70.6 & 90.9 \\
\rowcolor{bandDeepSeek}\icondeepseek~\texttt{deepseek-v4-pro} & 67.7 & 56.6 & 54.8 & 58.1 & 38.8 & 62.0 & 64.6 & 50.7 & 75.6 & 58.6 & 85.3 & 65.5 & 80.8 & 62.0 & 68.2 & 91.3 \\
\rowcolor{bandDeepSeek}\icondeepseek~\texttt{deepseek-v4-pro-0813} & 74.8 & 61.5 & 59.3 & 66.4 & 49.7 & 67.8 & 70.9 & 61.8 & 82.2 & 64.4 & 82.4 & 74.1 & 92.3 & 74.0 & 86.4 & 78.3 \\
\rowcolor{bandDeepSeek}\icondeepseek~\texttt{deepseek-v4-flash} & 55.9 & 42.2 & 38.0 & 41.5 & 31.2 & 41.5 & 54.4 & 44.1 & 72.6 & 54.0 & 67.6 & 62.1 & 88.5 & 46.0 & 59.1 & 60.9 \\
\midrule
\emph{median}, \env{sur} & 82.1 & 77.5 & 80.8 & 81.2 & 66.7 & 85.1 & 94.7 & 70.6 & 86.2 & 82.4 & 75.0 & 64.3 & 100.0 & 60.0 & 100.0 & 100.0 \\
\emph{median}, \env{cha} & 73.2 & 52.4 & 49.3 & 49.3 & 34.4 & 49.0 & 81.5 & 50.0 & 87.8 & 59.3 & 90.0 & 78.6 & 100.0 & 68.8 & 71.4 & 71.4 \\
\emph{median}, \env{web} & 70.2 & 61.9 & 54.1 & 68.1 & 52.9 & 63.8 & 63.6 & 71.0 & 86.2 & 83.7 & 87.5 & 76.7 & 95.2 & 70.8 & 70.0 & 100.0 \\
\emph{median}, \env{app} & 69.7 & 67.1 & 64.9 & 73.0 & 48.4 & 69.0 & 78.1 & 78.1 & 73.5 & 73.4 & 82.4 & 71.9 & 86.4 & 70.4 & 75.0 & 100.0 \\
\bottomrule
\end{tabular}
\end{table*}

Table~\ref{tab:bycategory} provides a granular breakdown of persona adherence across 16 trait categories grouped into five broad domains: Interests, Behavior, Values, Personality, and Communication/Expertise/Health; each cell pools single- and multi-attribute tasks, with a multi-attribute task counted once in every category its checks cover. Adherence varies markedly across domains and interacts strongly with the evaluation medium. Across models, health lifestyle, values and risk attitudes are the most stable, with median pass rates of at least 71.4\% on every surface (health reaches 100.0\% on survey, web, and app). In contrast, granular everyday preferences such as food, media and sports experience steep declines when moving from passive survey questions to interactive environments: median food adherence drops from 81.2\% on survey to 49.3\% on chat, media from 85.1\% to 49.0\%, and sports from 66.7\% to 34.4\%. In chat and web environments, models are readily swayed by off-persona defaults or persuasive counter-proposals when making concrete consumption decisions. Co-exposing several traits widens these gaps: the median arm passes 72.7\% of single-attribute sports trials but 44.4\% of multi-attribute ones, and 85.5\% versus 52.2\% for culture.

\subsection{Cross-Surface Consistency}
\label{sec:matched}

To assess whether adherence generalizes across environments, we match single-attribute scenarios by scenario identity, persona hash, and pinned trait set. Requiring complete, scored executions across all four interaction surfaces for all six OpenAI arms yields a matched cohort of 140 scenarios spanning 44 distinct personas (154 single-attribute scenarios in the benchmark are realized on all four surfaces). Holding the latent persona and operational errand fixed allows us to isolate the effect of interaction modality from persona sampling variance.

\begin{figure}[htbp]
\centering
\includegraphics[width=\linewidth]{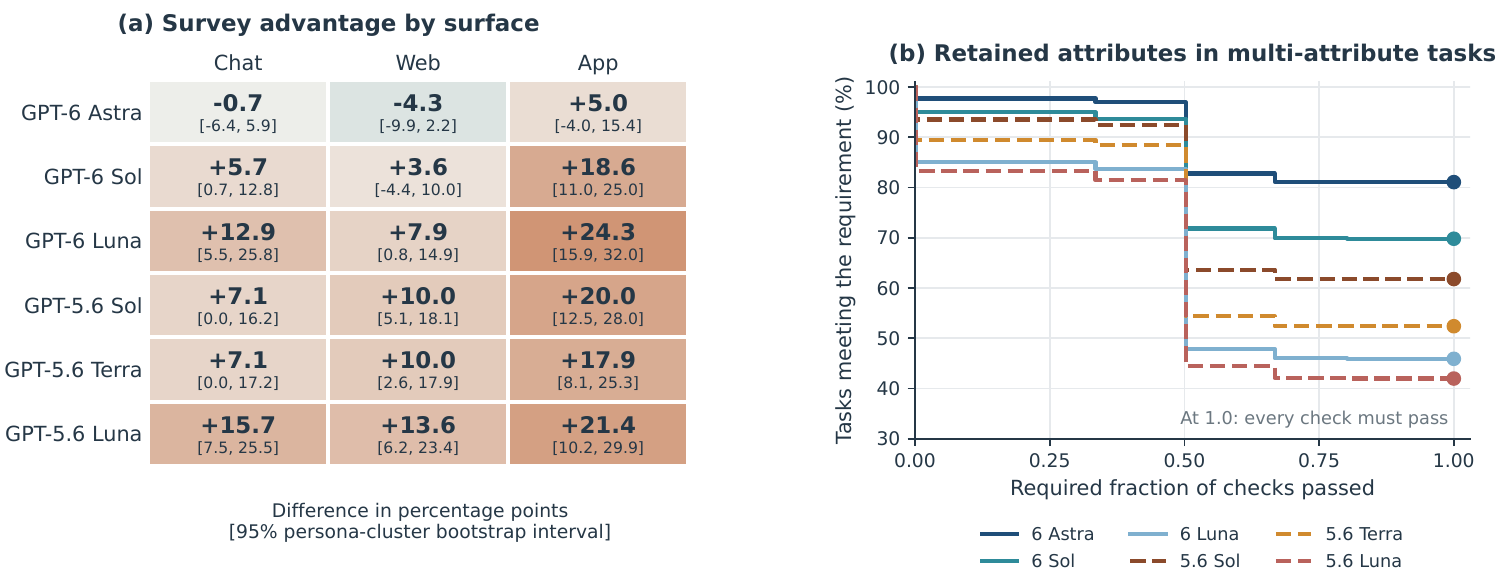}
\caption{\textbf{Surface gaps and retained attributes.}
(a) Survey minus chat, web, or app full-pass rate on 140 matched scenarios.
Warm cells favor survey; cool cells favor the interactive surface. Each cell
reports the paired difference and its 95\% interval from 10,000 bootstrap
resamples clustered by persona hash. (b) Percentage of 1,172 common
multi-attribute tasks meeting each normalized-credit threshold. Curves show
the empirical score distribution without smoothing; at a threshold of 1,
the value is the full-pass rate. Both panels use the same six OpenAI arms.}
\label{fig:analysis}
\end{figure}

Astra passes 80.0--89.3\% of scenarios on individual surfaces, but 64.3\% on all four. For GPT-5.6 Luna, survey adherence is 85.7\%, while all-four adherence is 37.9\%. Thus, a high score on one interface does not establish
that the same persona and scenario will pass on another. All-four adherence
is a stricter requirement than average adherence; its lower value alone does
not imply that surface failures are correlated.

Matching also qualifies the chat disadvantage. Astra scores 85.0\% on survey and 85.7\% on chat, a difference of $-0.7$ points (95\% interval $[-6.4,5.9]$). The survey advantage is 5.7 points for GPT-6 Sol ($[0.7,12.8]$) and 7.1 for both GPT-5.6 Sol and Terra (intervals reaching zero), but 12.9 for GPT-6 Luna ($[5.5,25.8]$) and 15.7 for GPT-5.6 Luna ($[7.5,25.5]$). The chat deficit therefore grows as models weaken rather than holding uniformly within matched single-attribute scenarios.

The app gap is more consistent (Figure~\ref{fig:analysis}a). Survey exceeds
app for all six arms: by 18.6 points for GPT-6 Sol ($[11.0,25.0]$), 24.3 for GPT-6 Luna ($[15.9,32.0]$), and 21.4 for GPT-5.6 Luna ($[10.2,29.9]$). Astra's smaller gap, 5.0 points, has an interval crossing zero ($[-4.0,15.4]$). Including recorded execution errors as failures expands the cohort to 149 scenarios and preserves the direction of every survey--app
comparison. These comparisons hold the persona and scenario fixed, while
execution and verification still differ by surface; they do not identify
interaction length as the cause of the gap.

\begin{figure}[tbp]
\centering
\includegraphics[width=\linewidth]{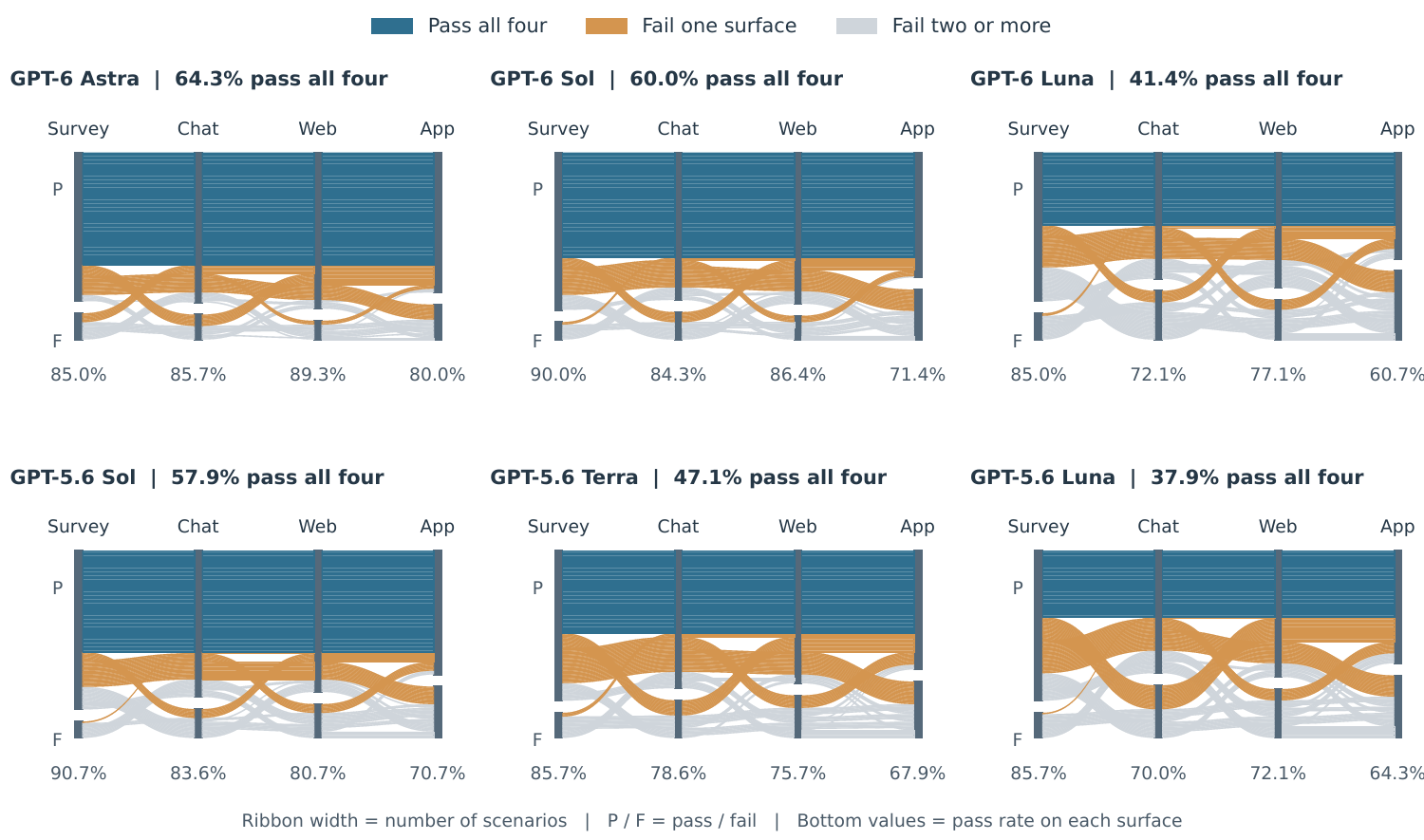}
\caption{\textbf{Matched outcomes across four surfaces.}
Each panel tracks the same 140 scenarios through their recorded survey,
chat, web, and app outcomes. At each surface, passing scenarios occupy the
upper block and failing scenarios the lower block. Ribbon widths are
proportional to scenario counts; colors indicate how many surfaces a
scenario fails across all four. Titles give all-four pass rates; bottom
values give marginal pass rates. Connections match scenario identities
across independent surface runs and do not represent a temporal sequence.}
\label{fig:transfer}
\end{figure}

\paragraph{Equal average scores can hide different outcomes.}
Astra passes 106 of the 140 scenarios on both survey and chat, yet 13 survey passes fail on chat and 14 survey failures pass on chat. The net gap is one scenario, but 27 scenarios (19.3\%) change outcome. The ribbons in
Figure~\ref{fig:transfer} distinguish stable outcomes from these offsetting
gains and losses across the four surfaces. Conditioning on survey passes,
17 of Astra's 119 passes fail on app (14.3\%); for GPT-5.6 Luna, 37 of 120 fail (30.8\%). App has the largest conditional failure rate
for every arm. These results quantify the limits of using a questionnaire
pass as evidence of adherence in an interactive setting. The different
model-specific denominators should be kept in mind when comparing rates.

\paragraph{Failures are often limited to one surface.}
Among scenarios failing on at least one surface, 43.9--62.0\% fail on exactly one (orange ribbons in Figure~\ref{fig:transfer}). Only 2--5 of the 140 scenarios per arm fail on all four. For GPT-5.6 Luna, 17 scenarios pass survey, chat, and web but fail app; Astra has 12 such scenarios. Thus, many failures
concern particular interface conditions rather than a trait that is missed
on every surface. Because the surface implementations and verifiers differ,
these patterns locate the disagreement without identifying its cause.

\subsection{Joint Trait Adherence}
\label{sec:composition}

We next compare the six arms on 1,172 shared multi-attribute tasks.
Figure~\ref{fig:analysis}b shows the fraction of tasks meeting each
normalized-credit threshold. Its right endpoints give full-pass rates; the
drop as the threshold increases shows how much adherence is partial.
Astra passes every check on 81.1\% of tasks, retains some checks on 16.6\%, and receives zero credit on 2.3\%. GPT-5.6 Luna has a lower full-pass rate, 42.0\%, but partial credit accounts for another 41.3\%. Across the six arms, 71.2--87.8\% of failed tasks retain at least one check.
Most observed failures therefore concern a subset of the measured traits.

This distinction matters when comparing models. GPT-6 Sol exceeds GPT-6 Luna by 23.9 points in full-pass rate (69.8\% versus 45.9\%), but by 16.9 points in mean normalized credit (82.5\% versus 65.6\%). The former
measures whether every requirement is met; the latter measures how much
adherence remains. Neither establishes an additional cost of combining
traits: even independent checks become harder to satisfy jointly as
their number increases. Such a claim requires matched single- and
multi-attribute versions of the same scenario. 

\subsection{Model Differences Within and Across Families}
\label{sec:cross-model}

\begin{figure}[tbp]
\centering
\includegraphics[width=.94\linewidth]{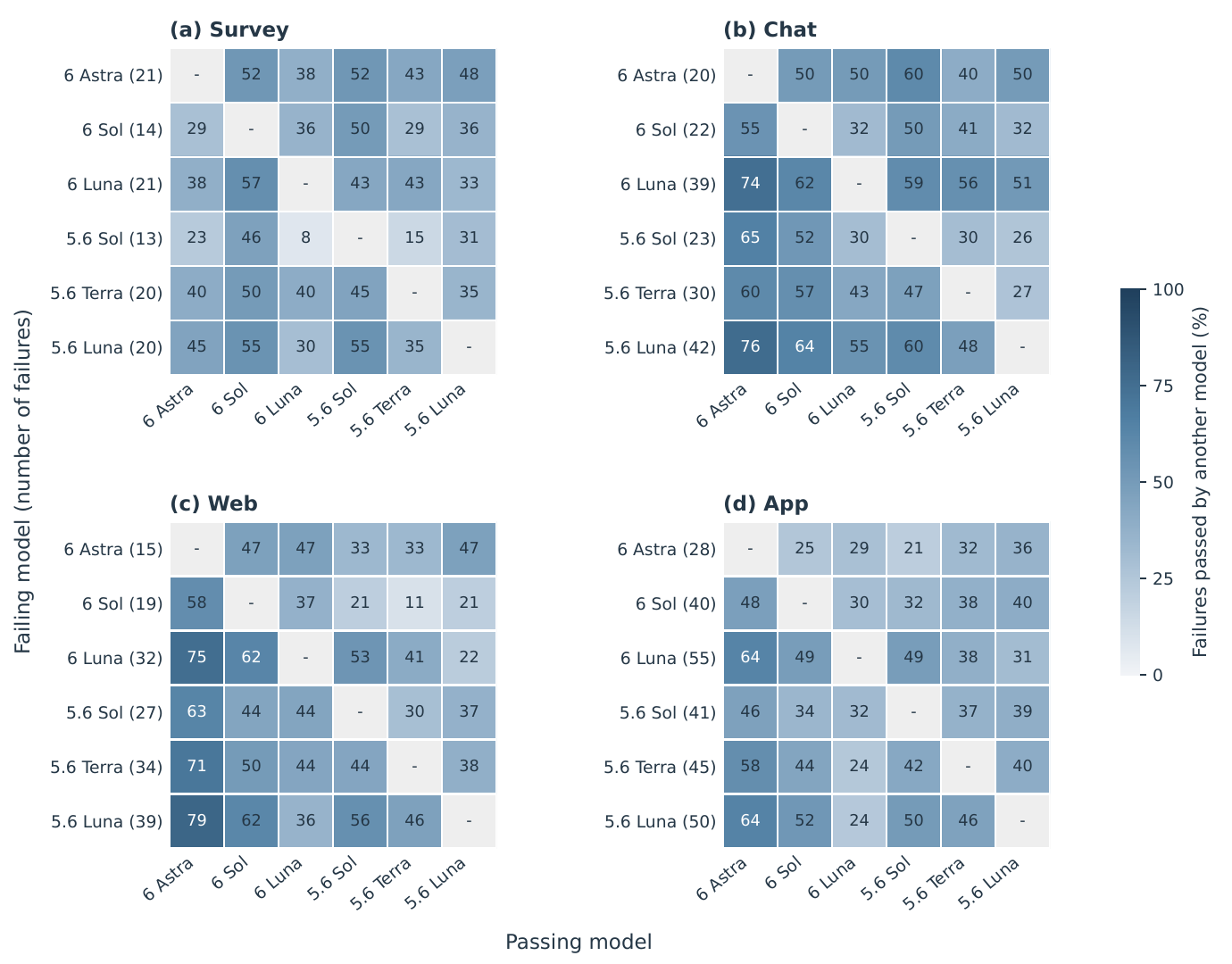}
\caption{\textbf{Cross-model differences on all four surfaces.}
Each panel uses the same six OpenAI arms and 140 matched scenarios.
Cells give the percentage of the row model's failed scenarios passed by
the column model, rounded to whole percentages. Parentheses give the row's
failure denominator; a high percentage can therefore represent few tasks.
The shared color scale permits comparison across surfaces. All labels refer
to GPT models. The eighteen-arm cross-vendor analysis is reported separately
in Figure~\ref{fig:cross-vendor}.}
\label{fig:failure-patterns}
\end{figure}

We separate two comparisons. First, the six OpenAI arms are compared on all
four surfaces using the same 140 scenario quartets as
Section~\ref{sec:matched}. This holds the model set and task composition
fixed across panels. Second, eighteen arms are compared on their larger common
chat/web cohort, where the evaluation environment is shared across vendors. Including
survey and app in the latter comparison would also introduce vendor-specific
interface implementations, so we report the two cohorts separately.

\paragraph{Model differences persist on every surface.}
Figure~\ref{fig:failure-patterns} reports the fraction of each row model's
failures passed by another model, separately for survey, chat, web, and app.
GPT-5.6 Luna passes 10 of Astra's 21 survey failures, 10 of its 20 chat failures, 7 of its 15 web failures, and 10 of its 28 app failures. Thus,
the overlap between models' success sets changes with the interface, and
no single ordering describes all recorded outcomes. The row denominators
are shown explicitly: the web percentage, for example, corresponds to 7 scenarios and should not be read as a repeated-run success probability.

\paragraph{Cross-vendor agreement in a unified evaluation setting.}
The eighteen-arm cohort contains the 1,086 tasks (383 chat and 703 web) scored by every arm that ran both surfaces on the shared chat and web harnesses under the \texttt{gpt-5.6-luna} judge. Astra has the highest full-pass rate on this cohort, 84.7\%, followed by Opus-5.5 at 84.0\% and Gemini-3.8 at 83.3\%. Of Astra's 166 failed tasks, Gemini-3.8 passes 89 (53.6\%), Opus-5.5 passes 81 (48.8\%), GPT-6 Sol passes 64 (38.6\%), and GPT-5.6 Luna passes 44 (26.5\%). Astra in turn passes 373 of the 495 tasks failed by GPT-5.6 Luna, while Luna passes 18 of Astra's 69 chat failures and 26 of its 97 web failures. Overall, all eighteen models pass 229 tasks and fail 12, while the remaining 845 tasks (77.8\%) have mixed outcomes; with more arms in the cohort, a mixed outcome becomes more likely by construction. Figure~\ref{fig:cross-vendor} in the appendix shows full pairwise comparisons and the empirical agreement distribution. These selected-run outcomes reflect substantial cross-vendor behavioral complementarity, though Section~\ref{sec:stability} shows that part of it is run-to-run variation, and unanimous failures also require inspection to separate shared behavioral limits from potential task-specific artifacts.

\subsection{Run-to-Run Stability}
\label{sec:stability}

\begin{table}[tbp]
\centering
\caption{\textbf{Run-to-run stability of the GPT-6 pair.} Three repeats with identical persona, task, prompt, effort, and judge.
Full-pass rate (\%) over scored trials per repeat; mean (SD) over the three repeats. $\kappa$: Cohen's kappa pooled over the three repeat pairs. \emph{Gap}: \texttt{gpt-6-sol}
minus \texttt{gpt-6-luna} on tasks scored in every repeat for both models, with a 95\% task-bootstrap interval.}
\label{tab:stability}
\small
\setlength{\tabcolsep}{4.0pt}
\begin{tabular}{@{}ll rrr r r r@{}}
\toprule
Surface & Model & R1 & R2 & R3 & Mean (SD) & $\kappa$ & Gap [95\% CI] \\
\midrule
\multirow{2}{*}{\env{survey} (502)} & \texttt{gpt-6-sol} & 85.4 & 87.4 & 85.4 & 86.1 (0.9) & 0.68 & \multirow{2}{*}{20.8 [16.1, 25.5]} \\
 & \texttt{gpt-6-luna} & 65.2 & 67.2 & 62.3 & 64.9 (2.0) & 0.54 &  \\
\addlinespace[2pt]
\multirow{2}{*}{\env{chat} (565)} & \texttt{gpt-6-sol} & 72.4 & 70.9 & 72.3 & 71.9 (0.7) & 0.46 & \multirow{2}{*}{8.9 [2.8, 15.0]} \\
 & \texttt{gpt-6-luna} & 60.0 & 60.6 & 59.7 & 60.1 (0.4) & 0.47 &  \\
\addlinespace[2pt]
\multirow{2}{*}{\env{web} (707)} & \texttt{gpt-6-sol} & 78.6 & 75.6 & 77.4 & 77.2 (1.2) & 0.71 & \multirow{2}{*}{19.7 [16.8, 22.6]} \\
 & \texttt{gpt-6-luna} & 57.7 & 56.1 & 56.6 & 56.8 (0.7) & 0.75 &  \\
\addlinespace[2pt]
\multirow{2}{*}{\env{app} (686)} & \texttt{gpt-6-sol} & 69.3 & 69.3 & 70.5 & 69.7 (0.6) & 0.73 & \multirow{2}{*}{13.5 [8.6, 18.7]} \\
 & \texttt{gpt-6-luna} & 55.2 & 55.5 & 56.1 & 55.6 (0.4) & 0.63 &  \\
\bottomrule
\end{tabular}
\end{table}

Every leaderboard cell comes from a single run per task. To measure how much a
rerun changes the result, we evaluate the GPT-6 pair (\texttt{gpt-6-sol} and
\texttt{gpt-6-luna}) three times on all four surfaces of the current task set
(502 survey, 565 chat, 707 web, and 686 app tasks), the same tasks as Table~\ref{tab:leaderboard}. Persona, task, prompt,
reasoning effort, and judge are identical across repeats, so the repeats differ
only through sampling in the evaluated model, the simulated conversational
partner, and the judge. Because survey and chat were re-run on the current
harness, their absolute rates differ from Table~\ref{tab:leaderboard}, whose
survey and chat cells come from the frozen-release runs.

\paragraph{Aggregate rates are stable.}
Table~\ref{tab:stability} shows that the standard deviation of a model's
surface-level pass rate across the three repeats is at most 2.0 points. Two
repeats of the same model on the same tasks differ by 1.1--2.3 points on
average and by at most 4.7. The sol--luna gap ranges from 8.9 to 20.8 points,
its 95\% interval excludes zero on every surface, and sol leads luna in all
twelve surface--repeat comparisons. Model differences of the size reported in
Table~\ref{tab:leaderboard} are therefore not artifacts of run-to-run variation.

\paragraph{Task-level agreement and dialogue dynamics.}
Across tasks scored in all three repeats, pairwise repeat agreement remains substantial, ranging from 74.3\% to 92.0\% (Cohen's $\kappa$ 0.46--0.75). Consistent with the stochasticity of an unscripted multi-turn conversational partner, \env{chat} exhibits lower agreement ($\kappa$ 0.46 and 0.47) than deterministic interface surfaces like \env{web} and \env{app} ($\kappa$ 0.63--0.75). A single run thus supports robust model-level comparisons at the resolution of a few points, though fine-grained task-level statements, such as which model passes another's failures (Section~\ref{sec:cross-model}), reflect empirical rates under natural dialogue variance rather than invariant properties of a task. Appendix~\ref{app:repeat-protocol} gives the protocol and the handling of execution errors.

\section{Discussion}
\label{sec:discussion}

\subsection{Decoupling Competence from Persona Fidelity}
\label{sec:competence-vs-fidelity}

Results across \textsc{APB} show that latent persona adherence operates largely orthogonally to general task competence. Standard post-training optimizes foundation models for helpfulness, technical accuracy, and unconstrained problem solving. Human simulation, by contrast, frequently requires operating under bounded rationality: respecting localized knowledge ceilings, subjective tastes, risk-averse heuristics, or declining mathematically optimal shortcuts. When a model exhibits capability intrusion (such as an elementary student persona spontaneously solving formal algebra), the error stems not from weak reasoning, but from an inability to suppress pre-trained capability in favor of assigned profile constraints.

Conversely, successful environmental execution does not guarantee persona compliance. In interactive browser and desktop environments, an agent may navigate multi-step interface operations (such as finding navigation paths, filling forms, and executing checkout flows) while selecting items that directly violate latent lifestyle or dietary constraints. For example, in \texttt{social-media-skeptic-web}, top-tier models like \texttt{gpt-6-astra} and \texttt{opus-5-5} complete the interface sequence to populate home-screen applications on a newly initialized device, yet select photo-sharing and public feed apps that violate the persona's latent skepticism toward social media. In contrast, arms such as \texttt{gpt-6-sol} pass by selecting local-only encrypted notes and a weather utility with minimal permissions. Here the failure is governed entirely by the selected content rather than interface execution competence. Benchmark competence metrics on suites such as SWE-bench or OSWorld capture whether an agent \emph{can} achieve an errand, but offer no guarantee regarding \emph{how} that errand is resolved relative to user profile specifications. Evaluating agentic systems therefore demands verifying behavioral boundaries alongside raw functional success.

\subsection{Cross-Surface Fragility and Attentional Load}
\label{sec:cross-surface-insights}

Our matched cohort evaluations reveal that conversational adherence provides an incomplete indicator of real-world steerability. Models that maintain high compliance in structured surveys or text-only dialogue frequently experience steep performance drops when deployed across interactive web and desktop software environments. An agent that readily declares vegetarian preferences in an interview may nonetheless select animal-derived ingredients when navigating a dense grocery delivery website. These cross-surface discrepancies indicate that persona conditioning is fragile, and that conversational role-play benchmarks overestimate an agent's true fidelity in practical computer-use environments.

Multi-attribute tasks introduce substantial attentional competition. While top-tier reasoning models like \texttt{gpt-6-astra} partially sustain multi-trait combinations, lighter and non-reasoning architectures suffer severe performance degradation under joint exposure, dropping by upwards of 25 percentage points. Holding multiple latent attributes in context without explicit per-turn reminders competes directly with the attentional capacity required for multi-step tool use and environmental navigation. Finally, our failure-overlap analyses (\S\ref{sec:cross-model}) demonstrate that no single model family uniformly subsumes another across all personas and surfaces. Different architectures display idiosyncratic behavioral blind spots, suggesting that robust user simulation in production workflows will benefit from heterogeneous model routing and surface-specialized ensembles rather than defaulting to any single monolithic model.

\section{Conclusion}
\label{sec:conclusion}

\mbox{AgentPersonaBench} evaluates latent persona adherence in frontier AI agents through unprompted, verified downstream behavior. Across 2,460 tasks, four distinct interaction surfaces, and 867 discrete traits, our empirical evaluation of 20 model arms demonstrates that persona fidelity cannot be inferred from conversational dialogue or aggregate benchmark competence. High adherence on one surface frequently coexists with failure on another, multi-attribute conditioning strains context allocation across all model families, and competing models exhibit pronounced behavioral complementarity. By evaluating concrete actions rather than conversational self-reports, \textsc{APB} establishes an empirical standard for testing whether autonomous agents can reliably simulate diverse human users in interactive digital environments.

\section*{Acknowledgments}
\label{sec:acknowledgments}

We thank Jierui Zuo, Ziwei Liu, Edmund Eric Gah, Yuanze Lin (University of
Oxford), Shenbo Xu (MIT / Kapnova Inc.), Zonglin Di (UCSC / UC Berkeley),
Heming Liu (University of Illinois Urbana-Champaign) and Yuxuan Cao (University of
Texas at Austin) for their contributions to the benchmark. We also thank our funders and partners for their support through compute resources,
model access, and research funding, which made the construction of AgentPersonaBench and the
evaluation runs reported here possible: OpenAI, Anthropic, Microsoft Azure,
Amazon Web Services, Meta, the MIT CSAIL Alliance, and the MIT Sandbox
Innovation Fund. The views and conclusions expressed in this paper are those of the
authors and do not necessarily reflect the positions of the supporting
organizations.

\clearpage
\nocite{apb2027}
\bibliographystyle{plainnat}
\bibliography{references}

\clearpage
\appendix
\etocdepthtag.toc{mtappendix}

\section*{Appendix Overview and Table of Contents}

This appendix provides author contributions and affiliations, extended
related literature, detailed per-category adherence breakdowns across all surfaces, the
reproducible analysis and audit protocol, benchmark limitations, future directions,
responsible use and artifact release, and ethical considerations
for \mbox{AgentPersonaBench} (\textsc{APB}).

\begingroup
\renewcommand{\contentsname}{}
\vspace{-1.5em}
\tableofcontents
\endgroup
\vspace{1.5em}
\clearpage
\section{Contributions and Affiliations}
\label{app:authors}

\organizerauthors, \contributorauthors, \advisoryauthors.

\subsection{Affiliations}
\label{app:author-affiliations}

\authoraffiliations

\clearpage
\section{Extended Related Work}
\label{app:related-work}

This section expands our discussion of prior literature across persona
agents, instruction following, steerability, and computer-use benchmarks.
Table~\ref{tab:related-work-comparison} summarizes the structural differences
between \textsc{APB} and representative benchmark paradigms.

\begin{table*}[t]
\centering
\small
\setlength{\tabcolsep}{4.5pt}
\renewcommand{\arraystretch}{1.2}
\begin{tabularx}{\textwidth}{
  >{\raggedright\arraybackslash\bfseries}p{2.7cm}
  >{\raggedright\arraybackslash}p{3.0cm}
  >{\raggedright\arraybackslash}p{3.2cm}
  >{\raggedright\arraybackslash}p{3.5cm}
  >{\raggedright\arraybackslash}X
}
\toprule
Benchmark Family & Representative Suites & Evaluation Target & Measurement Mode & Execution Substrate \\
\midrule
Conversational Role-Play &
  PersonaChat, RoleLLM, CharacterEval, Sotopia &
  Conversational tone, frame retention, dialogue voice &
  LLM-as-a-judge scoring of generated text; self-report &
  Multi-turn text dialogue \\
\midrule
Demographic Simulation &
  Out of One Many, Whose Opinions, Generative 1000 &
  Aggregate statistical alignment with human populations &
  Distributional distance on survey responses &
  Multiple-choice survey forms \\
\midrule
Instruction Following &
  IFEval, FollowBench, SysBench, SteerBench &
  Compliance with declared constraints or system rules &
  Deterministic verifiers on explicitly prompted instructions &
  Single- or multi-turn prompt-response \\
\midrule
Agentic Competence &
  WebArena, OSWorld, WorkArena, AppWorld &
  Functional task completion (e.g., book flight, write code) &
  Environment state diffs against task objectives &
  Web browser, native OS, terminal, or API environments \\
\midrule
\rowcolor{gray!12}
\textsc{APB} (Ours) &
  \textsc{APB} &
  Latent persona adherence under unprompted conditions &
  Deterministic verifiers on interface state, files, or actions &
  Survey, Chat, Web (Interactive), and App (Desktop GUI) \\
\bottomrule
\end{tabularx}
\caption{Comparison of \textsc{APB} with existing benchmark families. Unlike prior role-play benchmarks that evaluate superficial text generation and competence benchmarks that measure task completion, \textsc{APB} evaluates whether latent, unprompted persona attributes faithfully govern downstream behavioral decisions across four interaction surfaces of increasing realism.}
\label{tab:related-work-comparison}
\end{table*}

\subsection{Persona Agents, Role-Play, and User Simulation}
\label{app:related-user-sim}

Prior research on simulating humans with language models generally divides
into two lines of inquiry: character consistency in conversational role-play,
and synthetic population sampling in empirical social science. While both
directions offer valuable measurement tools, neither evaluates whether background
identity guides unprompted behavioral decisions.

\paragraph{Character consistency and conversational role-play.}
The first tradition investigates whether a language model can faithfully sustain an
assigned character persona. Foundational work centered on open-domain dialogue
conditioned on brief persona sentences \citep{zhang2018personachat}. Recent
advances extend this to literary and historical personas, evaluating whether
models sustain conversational voice, avoid breaking character, and accurately
retrieve biographical facts \citep{shao2023character, wang2023rolellm,
tu2024charactereval, chen2024persona, chen2024socialbench}. Subsequent frameworks
have assessed persona agents along multi-dimensional behavioral and social axes
\citep{samuel2024personagym}, administered standard psychometric inventories
(e.g., Big Five, MBTI) to role-played personas \citep{wang2024incharacter,
jiang2024personallm}, placed agents into dynamic social episodes to evaluate
multi-party interactions and negotiation \citep{zhou2024sotopia}, and systematically
measured how persona consistency degrades over extended multi-turn conversations
\citep{li2024measuring}. Comprehensive surveys chart the growth and taxonomies of
these role-playing systems \citep{tseng2024two}.

\paragraph{Synthetic cohorts and human substitution in social science.}
A complementary tradition deploys persona-conditioned models to stand in for human
participants in empirical social science. Researchers have examined whether models
can emulate representative survey respondents across demographic slices
\citep{argyle2023out, santurkar2023whose, durmus2024towards}, explored the
replication of classic behavioral economics and psychology experiments using
synthetic participant pools \citep{aher2023using, horton2023large}, and actively
debated the epistemic validity and representational fidelity of such substitutions
\citep{dillion2023can}. Large-scale interactive simulacra of human communities
\citep{park2023generative, park2024generative1000, li2026matraix} and extensive synthetic persona
registries \citep{ge2024scaling} provide rich demographic profiles for these
experiments. Concurrently, simulated users have increasingly been adopted as
automated, interactive evaluators for task-oriented agents in customer support and
e-commerce domains \citep{yao2024tau, barres2025tau2bench, zhang2024usimagent}.

\paragraph{The gap addressed by \textsc{APB}.}
Despite their breadth, both traditions leave a fundamental question unanswered.
Role-play benchmarks predominantly evaluate the \emph{surface} of a persona:
whether generated text ``sounds like'' the assigned character, typically evaluated
by LLM judges or surface perplexity. An agent composing eloquent Victorian prose
while placing meat in a vegetarian's grocery basket passes text-based role-play
rubrics but fails the core requirements of an acting proxy. Conversely,
human-substitution studies evaluate \emph{aggregate} population fidelity,
measuring whether marginal distribution statistics match demographic benchmarks.
However, aggregate calibration can be achieved by an unfaithful model whose
individual errors cancel out across cohorts. \textsc{APB} departs decisively from
both paradigms: it evaluates individual decisions under known, deterministic
ground truth. The target trait is \emph{never mentioned} in the task
prompt, requiring the agent to translate background identity into concrete,
observable action.

\subsection{Instruction Following, Steerability, and System-Prompt Adherence}
\label{app:related-instruction}

\paragraph{Instruction tuning and alignment.}
Instruction-tuned language models are optimized to follow explicit commands and
satisfy declarative constraints \citep{wei2022finetuned, wang2022super,
wang2023selfinstruct, ouyang2022training}. Reinforcement learning from human
feedback (RLHF) and direct preference optimization provide normative steering
signals \citep{christiano2017deep, stiennon2020learning, bai2022training,
rafailov2023direct, askell2021general}, while constitutional frameworks embed explicit
principles directly into self-refinement loops \citep{bai2022constitutional,
kundu2023specific}.

\paragraph{Constraint satisfaction and steerability benchmarks.}
A wide range of benchmarks evaluate how reliably models adhere to explicit
instructions. Verifiable constraint benchmarks assess whether models follow negative
constraints, formatting specifications, and length limitations
\citep{zhou2023instruction, jiang2024followbench, qin2024infobench,
zeng2024evaluating}, while others evaluate multi-turn or multilingual execution
\citep{he2024multiif, kwan2024mteval} or rank responses using pairwise LLM
judges \citep{zheng2023judging, chiang2024chatbot, dubois2024length}.
Parallel investigations probe whether models preserve system prompts under
adversarial user jailbreaks \citep{qin2024sysbench, mu2025closer,
wallace2024instruction}, and steerability suites measure how effectively a model
can be directed toward a specified persona, tone, or opinion \citep{dong2023steerlm,
chen2025steerbench}.

\paragraph{Explicit prompting vs.\ latent persona guidance.}
Instruction following differs fundamentally from latent persona adherence.
Instruction and steerability benchmarks explicitly prompt the expected behavior
(e.g., ``Respond concisely in pirate speech'' or ``Adopt the viewpoint of an environmentalist'').
Under direct prompting, models readily display sycophancy, mirroring stated constraints
\citep{sharma2023towards, perez2022discovering}, while their expressed principles
routinely diverge from concrete actions \citep{gu2025alignment}. In real-world applications,
a user profile is rarely an operational directive. Instead, an agent must independently
recognize when an underlying trait (such as budget consciousness, dietary restrictions,
or risk aversion) governs a routine action that makes no explicit mention of that trait.
\textsc{APB} isolates and benchmarks this implicit behavioral steering.

\subsection{Agentic Environments, Computer Use, and Competence Benchmarks}
\label{app:related-agentic}

\paragraph{Interactive environments and tool use.}
Recent agent benchmarks increasingly evaluate models operating within live,
interactive environments: executing browser interactions \citep{yao2022webshop,
zhou2024webarena, koh2024visualwebarena, drouin2024workarena, chezelles2024browsergym,
deng2023mind2web, he2024webvoyager}, interacting with operating systems and native
desktop interfaces \citep{xie2024osworld, bonatti2024windowsagentarena,
rawles2024androidworld, trivedi2024appworld}, utilizing tools and external APIs
\citep{qin2024toolllm, li2023apibank, patil2025bfcl}, and solving multi-modal
assistant challenges \citep{mialon2023gaia, liu2024agentbench}. Specialized suites
further benchmark complex real-world workflows, including software engineering
\citep{jimenez2024swebench, miserendino2025swelancer, merrill2026terminalbench},
machine-learning engineering \citep{chan2024mlebench}, enterprise CRM tasks
\citep{huang2025crmarena}, embodied simulation \citep{shridhar2021alfworld}, and
adversarial safety boundaries \citep{andriushchenko2024agentharm, zhang2024cybench}.

\paragraph{Competence vs.\ adherence: inverting the evaluation paradigm.}
These benchmarks fundamentally evaluate functional \emph{competence}: can the model
successfully navigate an operating system to configure a server, interact with an
e-commerce interface to buy a product, or patch a software bug? In contrast,
\textsc{APB} employs similar interactive substrates (web browser environments and desktop GUI applications operated via visual actions), but \emph{inverts} the evaluation goal. In \textsc{APB}, tasks are
engineered to be straightforward to execute from a functional standpoint, ensuring
that tool execution is a control variable rather than the principal barrier. The core
challenge is whether the model makes choices that conform to its assigned latent
persona when multiple functionally valid paths exist.

\paragraph{Methodological rigor and artifact verification.}
Recent critical analyses have shown that agent evaluations frequently suffer from
environment sensitivity, benchmark contamination, and trivial shortcut paths that
inflate perceived capability \citep{kapoor2024agents, zhu2025rigorous, kapoor2025hal,
raji2021everything}. \textsc{APB} safeguards against these failure modes through a
rigorous two-stage verification framework (\S\ref{sec:verification}): automated
contract audits verify that tasks cannot be solved blindly without persona
context, while expert human auditing ensures scenario naturalness and unambiguous
ground truth. By recording fine-grained execution traces and isolating interface
execution failures from genuine persona violations (\S\ref{sec:results}),
\textsc{APB} establishes a reproducible, discriminative standard for measuring persona
fidelity in interactive agents.

\clearpage
\section{Adherence by Category and Surface}
\label{app:bycategory}

Table~\ref{tab:bycategory-full} breaks the
per-category summary of Table~\ref{tab:bycategory} out by surface for the nineteen arms scored by \texttt{gpt-5.6-luna} (excluding \texttt{glm-5-3}, which used an alternative judge). As in Table~\ref{tab:bycategory}, each cell
pools single- and multi-attribute tasks, with a multi-attribute task counted once
in every category its checks cover. \emph{n} is the number of scored trials in a
category, given as the smallest count over the arms.
All arms use the evaluation set of Table~\ref{tab:leaderboard}; an arm without a valid run on a surface (marked N/A there) has no entries for it, and its \env{all} covers the remaining surfaces.

\definecolor{apbband1}{HTML}{FBF3E4}
\definecolor{apbband2}{HTML}{F7EFE6}
\definecolor{apbband3}{HTML}{F0EDF7}
\definecolor{apbband4}{HTML}{E9EFF9}
\definecolor{apbband5}{HTML}{E8F2EC}
{
\scriptsize
\setlength{\tabcolsep}{2.6pt}
\renewcommand{\arraystretch}{1.02}
\begin{longtable}{@{}ll|cccccc|cc|ccc|cc|ccc@{}}
\caption{\textbf{Adherence by dimension category and surface (\%).} Columns as in Table~\ref{tab:bycategory}; \env{all} is the full-pass rate over every surface; the last block is the median over all nineteen arms.}
\label{tab:bycategory-full} \\
\toprule
& & \multicolumn{6}{c|}{\textbf{Interests}} & \multicolumn{2}{c|}{\textbf{Behavior}} & \multicolumn{3}{c|}{\textbf{Values}} & \multicolumn{2}{c|}{\textbf{Person.}} & \multicolumn{3}{c}{\textbf{Comm./Exp./Health}} \\
\cmidrule(lr){3-8}\cmidrule(lr){9-10}\cmidrule(lr){11-13}\cmidrule(lr){14-15}\cmidrule(lr){16-18}
\textbf{Model} & & \env{cult} & \env{topic} & \env{hobby} & \env{food} & \env{sport} & \env{media} & \env{pref} & \env{habit} & \env{value} & \env{world} & \env{risk} & \env{big5} & \env{char} & \env{comm} & \env{exprt} & \env{health} \\
\emph{n} & & 127 & 325 & 221 & 229 & 188 & 460 & 79 & 67 & 135 & 87 & 34 & 58 & 26 & 50 & 22 & 23 \\
\midrule
\endfirsthead

\multicolumn{18}{l}{\footnotesize\textbf{Table~\ref{tab:bycategory-full}} (continued)} \\
\addlinespace[4pt]
\toprule
& & \multicolumn{6}{c|}{\textbf{Interests}} & \multicolumn{2}{c|}{\textbf{Behavior}} & \multicolumn{3}{c|}{\textbf{Values}} & \multicolumn{2}{c|}{\textbf{Person.}} & \multicolumn{3}{c}{\textbf{Comm./Exp./Health}} \\
\cmidrule(lr){3-8}\cmidrule(lr){9-10}\cmidrule(lr){11-13}\cmidrule(lr){14-15}\cmidrule(lr){16-18}
\textbf{Model} & & \env{cult} & \env{topic} & \env{hobby} & \env{food} & \env{sport} & \env{media} & \env{pref} & \env{habit} & \env{value} & \env{world} & \env{risk} & \env{big5} & \env{char} & \env{comm} & \env{exprt} & \env{health} \\
\midrule
\endhead

\midrule
\multicolumn{18}{r}{\emph{\scriptsize Continued on next page\dots}} \\
\endfoot

\bottomrule
\endlastfoot
\rowcolor{bandOpenAI}\iconopenai~\texttt{gpt-6-astra} & \env{sur} & 84.6 & 84.3 & 89.0 & 87.0 & 71.7 & 91.5 & 94.7 & 76.5 & 89.7 & 82.4 & 75.0 & 64.3 & 100.0 & 60.0 & 100.0 & 100.0 \\*
\rowcolor{bandOpenAI} & \env{cha} & 82.9 & 80.0 & 77.0 & 85.5 & 68.9 & 79.3 & 92.6 & 60.0 & 97.6 & 77.8 & 70.0 & 92.9 & 100.0 & 81.2 & 57.1 & 100.0 \\*
\rowcolor{bandOpenAI} & \env{web} & 74.5 & 86.4 & 87.8 & 87.9 & 77.9 & 90.2 & 87.9 & 90.3 & 86.2 & 83.7 & 93.8 & 86.7 & 90.5 & 75.0 & 70.0 & 100.0 \\*
\rowcolor{bandOpenAI} & \env{app} & 77.8 & 79.7 & 74.3 & 82.8 & 56.2 & 81.1 & 81.2 & 87.5 & 73.5 & 71.4 & 94.1 & 71.9 & 90.9 & 66.7 & 75.0 & 100.0 \\*
\rowcolor{bandOpenAI} & \env{all} & \textbf{79.7} & \textbf{82.6} & \textbf{82.1} & \textbf{85.8} & \textbf{68.8} & \textbf{85.5} & \textbf{88.3} & \textbf{81.0} & \textbf{84.7} & \textbf{78.7} & \textbf{86.3} & \textbf{78.9} & \textbf{91.7} & \textbf{71.4} & \textbf{73.5} & \textbf{100.0} \\
\midrule
\rowcolor{bandOpenAI}\iconopenai~\texttt{gpt-6-sol} & \env{sur} & 84.6 & 87.3 & 91.8 & 91.3 & 86.7 & 92.9 & 94.7 & 70.6 & 86.2 & 88.2 & 75.0 & 57.1 & 100.0 & 60.0 & 100.0 & 100.0 \\*
\rowcolor{bandOpenAI} & \env{cha} & 70.7 & 69.5 & 75.7 & 71.0 & 52.5 & 72.4 & 96.3 & 65.0 & 90.2 & 66.7 & 80.0 & 78.6 & 100.0 & 75.0 & 85.7 & 85.7 \\*
\rowcolor{bandOpenAI} & \env{web} & 72.3 & 72.9 & 63.5 & 80.2 & 60.3 & 74.7 & 63.6 & 80.6 & 89.2 & 86.0 & 87.5 & 86.7 & 100.0 & 87.5 & 70.0 & 100.0 \\*
\rowcolor{bandOpenAI} & \env{app} & 60.0 & 69.1 & 48.6 & 64.4 & 48.4 & 68.0 & 71.9 & 78.1 & 80.9 & 71.4 & 82.4 & 53.1 & 90.9 & 81.5 & 83.3 & 100.0 \\*
\rowcolor{bandOpenAI} & \env{all} & \textbf{71.5} & \textbf{74.3} & \textbf{70.1} & \textbf{76.3} & \textbf{61.7} & \textbf{76.5} & \textbf{79.3} & \textbf{75.0} & \textbf{86.2} & \textbf{77.9} & \textbf{82.4} & \textbf{68.9} & \textbf{95.8} & \textbf{79.2} & \textbf{82.4} & \textbf{97.0} \\
\midrule
\rowcolor{bandOpenAI}\iconopenai~\texttt{gpt-5-6-sol} & \env{sur} & 89.7 & 72.4 & 77.1 & 83.3 & 69.1 & 81.7 & 89.5 & 76.5 & 93.1 & 87.5 & 87.5 & 64.3 & 100.0 & 70.0 & 100.0 & 100.0 \\*
\rowcolor{bandOpenAI} & \env{cha} & 76.9 & 66.7 & 66.1 & 62.1 & 46.5 & 65.2 & 84.6 & 63.2 & 94.6 & 65.2 & 90.0 & 83.3 & 100.0 & 68.8 & 40.0 & 66.7 \\*
\rowcolor{bandOpenAI} & \env{web} & 68.1 & 64.4 & 54.1 & 70.3 & 52.9 & 66.1 & 69.7 & 71.0 & 84.6 & 83.7 & 87.5 & 86.7 & 95.2 & 75.0 & 60.0 & 87.5 \\*
\rowcolor{bandOpenAI} & \env{app} & 64.4 & 59.3 & 45.7 & 72.4 & 40.6 & 63.3 & 84.4 & 68.8 & 67.6 & 62.9 & 82.4 & 59.4 & 77.3 & 66.7 & 66.7 & 100.0 \\*
\rowcolor{bandOpenAI} & \env{all} & \textbf{74.1} & \textbf{65.3} & \textbf{60.5} & \textbf{72.2} & \textbf{52.2} & \textbf{68.6} & \textbf{80.9} & \textbf{69.7} & \textbf{81.9} & \textbf{74.4} & \textbf{86.3} & \textbf{72.7} & \textbf{87.5} & \textbf{70.1} & \textbf{65.6} & \textbf{90.6} \\
\midrule
\rowcolor{bandOpenAI}\iconopenai~\texttt{gpt-5-6-terra} & \env{sur} & 74.4 & 71.6 & 76.7 & 65.2 & 45.0 & 74.5 & 94.7 & 58.8 & 82.8 & 70.6 & 75.0 & 64.3 & 100.0 & 60.0 & 100.0 & 100.0 \\*
\rowcolor{bandOpenAI} & \env{cha} & 73.2 & 54.3 & 47.9 & 49.3 & 34.4 & 49.0 & 92.6 & 45.0 & 97.6 & 46.2 & 90.0 & 64.3 & 100.0 & 73.3 & 100.0 & 71.4 \\*
\rowcolor{bandOpenAI} & \env{web} & 61.7 & 56.8 & 43.2 & 53.8 & 30.9 & 57.5 & 57.6 & 58.1 & 87.7 & 81.4 & 93.8 & 70.0 & 100.0 & 83.3 & 60.0 & 75.0 \\*
\rowcolor{bandOpenAI} & \env{app} & 51.1 & 48.0 & 37.1 & 60.9 & 37.5 & 58.6 & 62.5 & 62.5 & 72.1 & 68.6 & 82.4 & 65.6 & 77.3 & 70.4 & 66.7 & 70.0 \\*
\rowcolor{bandOpenAI} & \env{all} & \textbf{64.5} & \textbf{57.1} & \textbf{51.4} & \textbf{57.3} & \textbf{36.8} & \textbf{59.6} & \textbf{73.9} & \textbf{57.0} & \textbf{83.7} & \textbf{68.6} & \textbf{86.3} & \textbf{66.7} & \textbf{89.6} & \textbf{73.7} & \textbf{76.5} & \textbf{78.8} \\
\midrule
\rowcolor{bandOpenAI}\iconopenai~\texttt{gpt-6-luna} & \env{sur} & 74.4 & 60.8 & 64.4 & 66.7 & 28.3 & 70.2 & 89.5 & 58.8 & 89.7 & 64.7 & 87.5 & 57.1 & 100.0 & 60.0 & 100.0 & 100.0 \\*
\rowcolor{bandOpenAI} & \env{cha} & 75.0 & 52.4 & 56.8 & 50.7 & 37.7 & 56.6 & 81.5 & 55.0 & 82.9 & 63.0 & 90.0 & 78.6 & 100.0 & 68.8 & 71.4 & 85.7 \\*
\rowcolor{bandOpenAI} & \env{web} & 55.3 & 44.1 & 29.7 & 47.3 & 27.9 & 39.7 & 54.5 & 67.7 & 86.2 & 69.8 & 87.5 & 66.7 & 95.2 & 70.8 & 60.0 & 87.5 \\*
\rowcolor{bandOpenAI} & \env{app} & 44.4 & 44.7 & 30.0 & 55.2 & 28.1 & 48.5 & 62.5 & 62.5 & 66.2 & 57.1 & 70.6 & 62.5 & 81.8 & 66.7 & 75.0 & 100.0 \\*
\rowcolor{bandOpenAI} & \env{all} & \textbf{61.4} & \textbf{50.0} & \textbf{45.4} & \textbf{54.4} & \textbf{30.4} & \textbf{52.8} & \textbf{69.4} & \textbf{62.0} & \textbf{79.3} & \textbf{63.9} & \textbf{82.4} & \textbf{65.6} & \textbf{89.6} & \textbf{67.5} & \textbf{73.5} & \textbf{93.9} \\
\midrule
\rowcolor{bandOpenAI}\iconopenai~\texttt{gpt-5-6-luna} & \env{sur} & 74.4 & 55.9 & 65.8 & 69.6 & 35.0 & 68.1 & 94.7 & 64.7 & 79.3 & 70.6 & 62.5 & 64.3 & 100.0 & 60.0 & 100.0 & 100.0 \\*
\rowcolor{bandOpenAI} & \env{cha} & 58.5 & 32.4 & 27.0 & 34.8 & 24.6 & 33.1 & 77.8 & 25.0 & 87.8 & 55.6 & 80.0 & 71.4 & 75.0 & 68.8 & 42.9 & 71.4 \\*
\rowcolor{bandOpenAI} & \env{web} & 66.0 & 39.8 & 35.1 & 53.8 & 25.0 & 42.0 & 51.5 & 67.7 & 81.5 & 55.8 & 87.5 & 70.0 & 95.2 & 66.7 & 60.0 & 87.5 \\*
\rowcolor{bandOpenAI} & \env{app} & 55.6 & 43.9 & 31.4 & 62.1 & 20.3 & 51.5 & 59.4 & 65.6 & 55.9 & 71.4 & 82.4 & 43.8 & 68.2 & 70.4 & 66.7 & 100.0 \\*
\rowcolor{bandOpenAI} & \env{all} & \textbf{63.4} & \textbf{42.9} & \textbf{39.9} & \textbf{55.4} & \textbf{26.1} & \textbf{48.3} & \textbf{67.6} & \textbf{58.0} & \textbf{73.9} & \textbf{62.3} & \textbf{80.4} & \textbf{60.0} & \textbf{81.2} & \textbf{67.5} & \textbf{64.7} & \textbf{90.9} \\
\midrule
\rowcolor{bandAnthropic}\iconclaude~\texttt{opus-5-5} & \env{sur} & 79.5 & 83.3 & 86.3 & 89.9 & 71.7 & 92.2 & 100.0 & 70.6 & 93.1 & 88.2 & 87.5 & 64.3 & 100.0 & 60.0 & 80.0 & 100.0 \\*
\rowcolor{bandAnthropic} & \env{cha} & 84.8 & 83.5 & 87.5 & 90.5 & 63.2 & 82.8 & 91.3 & 64.7 & 97.1 & 75.0 & 87.5 & 77.8 & 100.0 & 75.0 & 85.7 & 100.0 \\*
\rowcolor{bandAnthropic} & \env{web} & 76.6 & 87.3 & 86.5 & 89.0 & 72.1 & 90.2 & 84.8 & 87.1 & 93.8 & 83.7 & 87.5 & 80.0 & 85.7 & 62.5 & 70.0 & 100.0 \\*
\rowcolor{bandAnthropic} & \env{app} & 75.6 & 84.6 & 87.1 & 86.2 & 67.2 & 88.2 & 78.1 & 84.4 & 79.4 & 79.4 & 88.2 & 75.0 & 86.4 & 70.4 & 75.0 & 100.0 \\*
\rowcolor{bandAnthropic} & \env{all} & \textbf{78.7} & \textbf{84.8} & \textbf{86.9} & \textbf{88.7} & \textbf{68.7} & \textbf{88.7} & \textbf{86.9} & \textbf{79.4} & \textbf{89.3} & \textbf{81.4} & \textbf{87.8} & \textbf{75.3} & \textbf{87.5} & \textbf{67.5} & \textbf{76.5} & \textbf{100.0} \\
\midrule
\rowcolor{bandAnthropic}\iconclaude~\texttt{opus-5} & \env{sur} & 87.2 & 84.3 & 90.4 & 79.7 & 71.7 & 87.9 & 94.7 & 82.4 & 89.7 & 82.4 & 75.0 & 78.6 & 100.0 & 70.0 & 80.0 & 100.0 \\*
\rowcolor{bandAnthropic} & \env{cha} & 73.2 & 57.3 & 61.6 & 50.7 & 37.3 & 47.5 & 85.2 & 65.0 & 95.1 & 61.5 & 60.0 & 78.6 & 100.0 & 87.5 & 85.7 & 85.7 \\*
\rowcolor{bandAnthropic} & \env{web} & 70.2 & 71.2 & 70.3 & 74.7 & 54.4 & 71.3 & 69.7 & 74.2 & 93.8 & 83.7 & 93.8 & 86.7 & 95.2 & 75.0 & 70.0 & 100.0 \\*
\rowcolor{bandAnthropic} & \env{app} & 80.0 & 71.5 & 74.3 & 75.9 & 50.0 & 74.6 & 81.2 & 81.2 & 85.3 & 80.0 & 88.2 & 71.9 & 95.5 & 74.1 & 83.3 & 100.0 \\*
\rowcolor{bandAnthropic} & \env{all} & \textbf{77.3} & \textbf{71.1} & \textbf{74.1} & \textbf{71.0} & \textbf{53.4} & \textbf{70.6} & \textbf{81.1} & \textbf{76.0} & \textbf{90.6} & \textbf{77.7} & \textbf{82.4} & \textbf{78.9} & \textbf{95.8} & \textbf{76.6} & \textbf{79.4} & \textbf{97.0} \\
\midrule
\rowcolor{bandAnthropic}\iconclaude~\texttt{opus-4-8} & \env{sur} & 82.1 & 81.4 & 80.8 & 76.8 & 68.3 & 85.1 & 94.7 & 70.6 & 89.7 & 88.2 & 75.0 & 85.7 & 100.0 & 70.0 & 100.0 & 100.0 \\*
\rowcolor{bandAnthropic} & \env{cha} & 73.2 & 63.8 & 60.8 & 53.6 & 44.3 & 68.3 & 81.5 & 40.0 & 87.8 & 59.3 & 70.0 & 78.6 & 100.0 & 68.8 & 71.4 & 71.4 \\*
\rowcolor{bandAnthropic} & \env{web} & 78.7 & 67.8 & 58.1 & 60.4 & 52.9 & 59.2 & 57.6 & 80.6 & 95.4 & 83.7 & 87.5 & 76.7 & 100.0 & 66.7 & 50.0 & 100.0 \\*
\rowcolor{bandAnthropic} & \env{app} & 77.8 & 68.3 & 75.7 & 73.6 & 62.5 & 71.0 & 81.2 & 78.1 & 86.8 & 88.6 & 88.2 & 75.0 & 90.9 & 70.4 & 75.0 & 100.0 \\*
\rowcolor{bandAnthropic} & \env{all} & \textbf{77.9} & \textbf{70.1} & \textbf{68.7} & \textbf{66.1} & \textbf{56.9} & \textbf{70.3} & \textbf{76.6} & \textbf{70.0} & \textbf{90.1} & \textbf{80.3} & \textbf{82.4} & \textbf{77.8} & \textbf{95.8} & \textbf{68.8} & \textbf{70.6} & \textbf{93.9} \\
\midrule
\rowcolor{bandGoogle}\icongemini~\texttt{gemini-3-8-flash} & \env{sur} & 97.4 & 78.2 & 88.6 & 91.0 & 88.1 & 92.9 & 100.0 & 88.2 & 89.3 & 100.0 & 87.5 & 69.2 & 100.0 & 80.0 & 100.0 & 100.0 \\*
\rowcolor{bandGoogle} & \env{cha} & 82.9 & 70.6 & 76.4 & 86.4 & 69.5 & 82.5 & 92.6 & 95.0 & 95.1 & 81.5 & 90.0 & 100.0 & 100.0 & 87.5 & 71.4 & 100.0 \\*
\rowcolor{bandGoogle} & \env{web} & 85.1 & 80.5 & 86.5 & 79.1 & 79.4 & 83.9 & 84.8 & 80.6 & 90.8 & 90.7 & 100.0 & 83.3 & 95.2 & 70.8 & 70.0 & 100.0 \\*
\rowcolor{bandGoogle} & \env{app} & 86.7 & 84.6 & 85.7 & 83.9 & 84.4 & 87.6 & 96.9 & 87.5 & 82.4 & 74.3 & 88.2 & 87.5 & 95.5 & 81.5 & 75.0 & 100.0 \\*
\rowcolor{bandGoogle} & \env{all} & \textbf{87.8} & \textbf{78.8} & \textbf{84.3} & \textbf{84.6} & \textbf{80.4} & \textbf{86.6} & \textbf{92.8} & \textbf{87.0} & \textbf{88.6} & \textbf{85.2} & \textbf{92.2} & \textbf{85.4} & \textbf{95.8} & \textbf{79.2} & \textbf{76.5} & \textbf{100.0} \\
\midrule
\rowcolor{bandGoogle}\icongemini~\texttt{gemini-3-7-flash} & \env{sur} & 97.4 & 77.5 & 89.0 & 88.4 & 81.7 & 91.5 & 100.0 & 82.4 & 86.2 & 94.1 & 75.0 & 64.3 & 100.0 & 80.0 & 100.0 & 100.0 \\*
\rowcolor{bandGoogle} & \env{cha} & 80.5 & 69.5 & 66.2 & 84.1 & 68.9 & 77.2 & 92.6 & 75.0 & 100.0 & 85.2 & 90.0 & 85.7 & 75.0 & 81.2 & 85.7 & 85.7 \\*
\rowcolor{bandGoogle} & \env{web} & 89.4 & 78.8 & 79.7 & 81.3 & 76.5 & 79.9 & 87.9 & 77.4 & 93.8 & 93.0 & 81.2 & 80.0 & 100.0 & 70.8 & 60.0 & 100.0 \\*
\rowcolor{bandGoogle} & \env{app} & 84.4 & 82.1 & 77.1 & 85.1 & 78.1 & 85.2 & 90.6 & 84.4 & 83.8 & 82.9 & 82.4 & 78.1 & 95.5 & 70.4 & 75.0 & 100.0 \\*
\rowcolor{bandGoogle} & \env{all} & \textbf{87.8} & \textbf{77.2} & \textbf{78.0} & \textbf{84.5} & \textbf{76.3} & \textbf{83.3} & \textbf{91.9} & \textbf{80.0} & \textbf{90.6} & \textbf{88.5} & \textbf{82.4} & \textbf{77.8} & \textbf{95.8} & \textbf{74.0} & \textbf{76.5} & \textbf{97.0} \\
\midrule
\rowcolor{bandXAI}\icongrok~\texttt{grok-4-6} & \env{sur} & 92.3 & 78.4 & 87.7 & 88.4 & 66.7 & 89.4 & 100.0 & 76.5 & 86.2 & 76.5 & 87.5 & 71.4 & 100.0 & 60.0 & 100.0 & 100.0 \\*
\rowcolor{bandXAI} & \env{cha} & 74.4 & 27.9 & 31.3 & 40.3 & 20.7 & 29.9 & 73.1 & 55.0 & 90.0 & 65.2 & 100.0 & 71.4 & 75.0 & 87.5 & 83.3 & 100.0 \\*
\rowcolor{bandXAI} & \env{web} & 68.1 & 65.0 & 60.8 & 71.4 & 48.5 & 69.0 & 57.6 & 74.2 & 84.6 & 62.8 & 87.5 & 70.0 & 95.2 & 83.3 & 70.0 & 100.0 \\*
\rowcolor{bandXAI} & \env{app} & 66.7 & 66.1 & 62.7 & 73.8 & 45.2 & 68.4 & 78.1 & 80.6 & 73.5 & 73.5 & 88.2 & 74.2 & 86.4 & 59.3 & 91.7 & 90.0 \\*
\rowcolor{bandXAI} & \env{all} & \textbf{74.7} & \textbf{59.6} & \textbf{61.2} & \textbf{69.1} & \textbf{45.6} & \textbf{64.9} & \textbf{74.5} & \textbf{72.7} & \textbf{82.2} & \textbf{68.4} & \textbf{90.2} & \textbf{71.9} & \textbf{89.6} & \textbf{72.7} & \textbf{84.8} & \textbf{97.0} \\
\midrule
\rowcolor{bandGLM}\iconglm~\texttt{glm-5-3-flash} & \env{sur} & 82.1 & 66.7 & 68.5 & 71.0 & 56.7 & 74.5 & 73.7 & 64.7 & 82.8 & 76.5 & 75.0 & 64.3 & 100.0 & 50.0 & 100.0 & 75.0 \\*
\rowcolor{bandGLM} & \env{cha} & 58.5 & 36.2 & 32.4 & 34.8 & 30.0 & 34.5 & 66.7 & 35.0 & 75.6 & 48.1 & 90.0 & 78.6 & 100.0 & 68.8 & 28.6 & 57.1 \\*
\rowcolor{bandGLM} & \env{web} & 70.2 & 58.5 & 54.1 & 52.7 & 29.4 & 55.2 & 63.6 & 71.0 & 83.1 & 72.1 & 87.5 & 73.3 & 90.5 & 66.7 & 60.0 & 87.5 \\*
\rowcolor{bandGLM} & \env{app} & 66.7 & 56.1 & 58.6 & 66.7 & 45.3 & 59.8 & 78.1 & 78.1 & 69.1 & 68.6 & 82.4 & 71.9 & 81.8 & 63.0 & 75.0 & 90.0 \\*
\rowcolor{bandGLM} & \env{all} & \textbf{69.2} & \textbf{54.5} & \textbf{53.3} & \textbf{56.6} & \textbf{40.1} & \textbf{56.0} & \textbf{70.3} & \textbf{65.0} & \textbf{76.8} & \textbf{66.4} & \textbf{84.3} & \textbf{72.2} & \textbf{87.5} & \textbf{63.6} & \textbf{63.6} & \textbf{78.8} \\
\midrule
\rowcolor{bandQwen}\iconqwen~\texttt{qwen-3-8-27b} & \env{sur} & 79.5 & 73.5 & 71.2 & 69.6 & 58.3 & 76.6 & 94.7 & 58.8 & 79.3 & 82.4 & 62.5 & 50.0 & 100.0 & 70.0 & 75.0 & 87.5 \\*
\rowcolor{bandQwen} & \env{cha} & 56.1 & 33.3 & 39.2 & 38.2 & 13.1 & 37.5 & 74.1 & 30.0 & 75.6 & 51.9 & 90.0 & 50.0 & 100.0 & 81.2 & 57.1 & 85.7 \\*
\rowcolor{bandQwen} & \env{web} & 63.8 & 60.2 & 51.4 & 62.6 & 58.8 & 63.8 & 63.6 & 67.7 & 80.0 & 83.7 & 87.5 & 73.3 & 81.0 & 62.5 & 60.0 & 100.0 \\*
\rowcolor{bandQwen} & \env{app} & 57.8 & 58.1 & 42.6 & 49.3 & 40.0 & 57.7 & 71.9 & 71.9 & 69.0 & 73.3 & 71.4 & 75.0 & 81.2 & 59.1 & 72.7 & 80.0 \\*
\rowcolor{bandQwen} & \env{all} & \textbf{64.0} & \textbf{56.3} & \textbf{51.2} & \textbf{55.5} & \textbf{43.1} & \textbf{59.0} & \textbf{73.9} & \textbf{60.0} & \textbf{75.6} & \textbf{73.5} & \textbf{79.2} & \textbf{66.7} & \textbf{83.3} & \textbf{66.7} & \textbf{65.6} & \textbf{87.9} \\
\midrule
\rowcolor{bandQwen}\iconqwen~\texttt{qwen-3-8-flash} & \env{sur} & 74.4 & 66.7 & 76.7 & 79.7 & 60.0 & 75.2 & 89.5 & 52.9 & 79.3 & 70.6 & 75.0 & 71.4 & 100.0 & 60.0 & 100.0 & 100.0 \\*
\rowcolor{bandQwen} & \env{cha} & 68.3 & 45.0 & 49.3 & 44.8 & 29.8 & 45.8 & 85.2 & 35.0 & 87.5 & 50.0 & 90.0 & 57.1 & 100.0 & 62.5 & 71.4 & 57.1 \\*
\rowcolor{bandQwen} & \env{web} & 72.3 & 60.2 & 52.7 & 60.4 & 38.2 & 61.5 & 60.6 & 61.3 & 84.6 & 72.1 & 87.5 & 80.0 & 90.5 & 75.0 & 66.7 & 87.5 \\*
\rowcolor{bandQwen} & \env{app} & 72.7 & 68.1 & 67.1 & 75.3 & 48.4 & 69.6 & 77.4 & 80.6 & 74.6 & 82.9 & 88.2 & 71.9 & 86.4 & 73.1 & 83.3 & 80.0 \\*
\rowcolor{bandQwen} & \env{all} & \textbf{71.9} & \textbf{60.4} & \textbf{61.5} & \textbf{65.4} & \textbf{44.1} & \textbf{63.2} & \textbf{76.4} & \textbf{60.6} & \textbf{81.1} & \textbf{70.2} & \textbf{86.3} & \textbf{72.2} & \textbf{89.6} & \textbf{69.7} & \textbf{78.1} & \textbf{81.8} \\
\midrule
\rowcolor{bandMoonshot}\iconkimi~\texttt{kimi-k3} & \env{sur} & 82.1 & 78.2 & 75.3 & 81.2 & 66.7 & 82.1 & 94.7 & 76.5 & 86.2 & 70.6 & 87.5 & 71.4 & 100.0 & 70.0 & 80.0 & 100.0 \\*
\rowcolor{bandMoonshot} & \env{cha} & 75.6 & 44.8 & 36.5 & 49.3 & 31.1 & 51.0 & 74.1 & 35.0 & 85.4 & 55.6 & 70.0 & 78.6 & 100.0 & 62.5 & 57.1 & 71.4 \\*
\rowcolor{bandMoonshot} & \env{web} & 74.5 & 61.9 & 63.5 & 70.3 & 58.8 & 71.3 & 69.7 & 77.4 & 90.8 & 86.0 & 93.8 & 83.3 & 90.5 & 79.2 & 70.0 & 100.0 \\*
\rowcolor{bandMoonshot} & \env{app} & 75.0 & 65.3 & 71.0 & 69.8 & 62.5 & 74.7 & 78.1 & 71.9 & 72.6 & 73.5 & 82.4 & 69.0 & 77.8 & 70.4 & 75.0 & 90.0 \\*
\rowcolor{bandMoonshot} & \env{all} & \textbf{76.6} & \textbf{62.5} & \textbf{61.4} & \textbf{67.9} & \textbf{54.9} & \textbf{69.9} & \textbf{77.5} & \textbf{67.0} & \textbf{83.2} & \textbf{73.6} & \textbf{84.3} & \textbf{75.9} & \textbf{86.4} & \textbf{71.4} & \textbf{70.6} & \textbf{90.9} \\
\midrule
\rowcolor{bandDeepSeek}\icondeepseek~\texttt{deepseek-v4-pro} & \env{sur} & 74.4 & 74.5 & 80.8 & 84.1 & 58.3 & 87.9 & 89.5 & 47.1 & 75.9 & 82.4 & 75.0 & 64.3 & 100.0 & 50.0 & 60.0 & 100.0 \\*
\rowcolor{bandDeepSeek} & \env{cha} & 61.0 & 41.9 & 33.8 & 44.9 & 27.9 & 46.9 & 59.3 & 45.0 & 61.0 & 40.7 & 80.0 & 42.9 & 25.0 & 62.5 & 71.4 & 71.4 \\*
\rowcolor{bandDeepSeek} & \env{web} & 68.1 & 54.2 & 50.0 & 48.4 & 31.3 & 53.4 & 54.5 & 56.7 & 84.6 & 60.5 & 93.8 & 76.7 & 90.5 & 66.7 & 70.0 & 100.0 \\*
\rowcolor{bandDeepSeek} & \env{app} & -- & -- & -- & -- & -- & -- & -- & -- & -- & -- & -- & -- & -- & -- & -- & -- \\*
\rowcolor{bandDeepSeek} & \env{all} & \textbf{67.7} & \textbf{56.6} & \textbf{54.8} & \textbf{58.1} & \textbf{38.8} & \textbf{62.0} & \textbf{64.6} & \textbf{50.7} & \textbf{75.6} & \textbf{58.6} & \textbf{85.3} & \textbf{65.5} & \textbf{80.8} & \textbf{62.0} & \textbf{68.2} & \textbf{91.3} \\
\midrule
\rowcolor{bandDeepSeek}\icondeepseek~\texttt{deepseek-v4-pro-0813} & \env{sur} & 87.2 & 80.4 & 89.0 & 85.5 & 75.0 & 95.0 & 94.7 & 70.6 & 89.7 & 82.4 & 75.0 & 85.7 & 100.0 & 80.0 & 100.0 & 87.5 \\*
\rowcolor{bandDeepSeek} & \env{cha} & 63.4 & 46.7 & 37.8 & 44.9 & 32.8 & 46.2 & 63.0 & 50.0 & 75.6 & 40.7 & 90.0 & 64.3 & 75.0 & 62.5 & 100.0 & 57.1 \\*
\rowcolor{bandDeepSeek} & \env{web} & 74.5 & 58.5 & 51.4 & 68.1 & 42.6 & 63.8 & 63.6 & 64.5 & 83.1 & 72.1 & 81.2 & 73.3 & 95.2 & 79.2 & 70.0 & 87.5 \\*
\rowcolor{bandDeepSeek} & \env{app} & -- & -- & -- & -- & -- & -- & -- & -- & -- & -- & -- & -- & -- & -- & -- & -- \\*
\rowcolor{bandDeepSeek} & \env{all} & \textbf{74.8} & \textbf{61.5} & \textbf{59.3} & \textbf{66.4} & \textbf{49.7} & \textbf{67.8} & \textbf{70.9} & \textbf{61.8} & \textbf{82.2} & \textbf{64.4} & \textbf{82.4} & \textbf{74.1} & \textbf{92.3} & \textbf{74.0} & \textbf{86.4} & \textbf{78.3} \\
\midrule
\rowcolor{bandDeepSeek}\icondeepseek~\texttt{deepseek-v4-flash} & \env{sur} & 61.5 & 63.7 & 63.0 & 75.4 & 51.7 & 74.5 & 78.9 & 35.3 & 86.2 & 64.7 & 62.5 & 50.0 & 100.0 & 30.0 & 80.0 & 87.5 \\*
\rowcolor{bandDeepSeek} & \env{cha} & 53.7 & 34.3 & 32.4 & 20.3 & 23.0 & 33.1 & 48.1 & 30.0 & 61.0 & 51.9 & 70.0 & 71.4 & 100.0 & 31.2 & 28.6 & 42.9 \\*
\rowcolor{bandDeepSeek} & \env{web} & 53.2 & 30.5 & 18.9 & 31.9 & 20.6 & 21.8 & 45.5 & 58.1 & 73.8 & 51.2 & 68.8 & 63.3 & 85.7 & 62.5 & 70.0 & 50.0 \\*
\rowcolor{bandDeepSeek} & \env{app} & -- & -- & -- & -- & -- & -- & -- & -- & -- & -- & -- & -- & -- & -- & -- & -- \\*
\rowcolor{bandDeepSeek} & \env{all} & \textbf{55.9} & \textbf{42.2} & \textbf{38.0} & \textbf{41.5} & \textbf{31.2} & \textbf{41.5} & \textbf{54.4} & \textbf{44.1} & \textbf{72.6} & \textbf{54.0} & \textbf{67.6} & \textbf{62.1} & \textbf{88.5} & \textbf{46.0} & \textbf{59.1} & \textbf{60.9} \\
\midrule
\emph{median, all arms} & \env{sur} & 82.1 & 77.5 & 80.8 & 81.2 & 66.7 & 85.1 & 94.7 & 70.6 & 86.2 & 82.4 & 75.0 & 64.3 & 100.0 & 60.0 & 100.0 & 100.0 \\
 & \env{cha} & 73.2 & 52.4 & 49.3 & 49.3 & 34.4 & 49.0 & 81.5 & 50.0 & 87.8 & 59.3 & 90.0 & 78.6 & 100.0 & 68.8 & 71.4 & 71.4 \\
 & \env{web} & 70.2 & 61.9 & 54.1 & 68.1 & 52.9 & 63.8 & 63.6 & 71.0 & 86.2 & 83.7 & 87.5 & 76.7 & 95.2 & 70.8 & 70.0 & 100.0 \\
 & \env{app} & 69.7 & 67.1 & 64.9 & 73.0 & 48.4 & 69.0 & 78.1 & 78.1 & 73.5 & 73.4 & 82.4 & 71.9 & 86.4 & 70.4 & 75.0 & 100.0 \\
\end{longtable}
}

\clearpage
\section{Reproducible Analysis Protocol}
\label{app:analysis-audit}

The evaluation and diagnostic protocols in Section~\ref{sec:results} operate on the official \textsc{APB} benchmark release and recorded model execution transcripts. All analyses evaluate recorded runtime artifacts directly, issuing no additional model generation calls and preserving original execution trajectories without altering model outputs. The main leaderboard and category breakdown tables reflect full release evaluations; detailed diagnostic analyses employ strictly matched cohorts to isolate interaction modalities and model architectures.

\paragraph{Task specification and scoring metrics.}
Each task is identified by its canonical path, single- or multi-attribute classification, and interaction surface. For a task declaring $K_t$ behavioral checks, we compute both full pass, $\mathbf{1}[s_t = K_t]$ (requiring all latent behavioral criteria to hold simultaneously), and normalized credit, $s_t / K_t$, across scored trials. Outcomes are determined strictly by verified verifier findings (rule-based assertions or rubric-guided evaluations), without inferring post-hoc credit.

\paragraph{Matched cohort construction and uncertainty estimation.}
To assess cross-surface transfer without confounding from persona sampling or task difficulty, single-attribute scenarios are matched across surfaces by shared scenario template, persona profile hash, and pinned trait specification. The primary four-surface matched cohort contains 140 scenarios across 44 distinct personas evaluated across all four surfaces by the six OpenAI arms. Pointwise 95\% confidence intervals are estimated via 10,000 bootstrap resamples clustered by persona hash, capturing behavioral variation across the sampled persona distribution. For multi-attribute analyses, the matched cohort contains 1,172 tasks evaluated across models under identical trait co-exposure conditions.

\paragraph{Execution error accounting.}
Reported benchmark adherence rates are conditional on valid completed executions. Infrastructure-level execution errors (such as transient environment timeouts or interface interaction anomalies) are tracked separately. To assess the sensitivity of leaderboard rankings to execution reliability, we conduct an error-inclusive audit where all non-completed attempts are assigned zero credit (\S\ref{sec:robustness}), confirming that relative model rankings and performance separations remain stable.

\subsection{Evaluation Coverage and Standardized Environments}
\label{sec:report-coverage}

We evaluate 20 foundation model arms across eight model families (Table~\ref{tab:leaderboard}) across the 2,460 benchmark tasks (502 survey, 565 chat, 707 web, and 686 app). Each model is packaged as a standardized, versioned arm recording its checkpoint, decoding parameters (medium reasoning budget where supported), and interaction wrapper. All reported scores are evaluated against frozen release benchmark environments, ensuring byte-level reproducibility across trials.

Due to input modality constraints, models lacking native vision or desktop GUI capabilities (e.g., text-only architectures including the DeepSeek series and \texttt{glm-5-3}) participate on the text surfaces (\env{survey}, \env{chat}, and \env{web}), with desktop interaction (\env{app}) designated as N/A; more broadly, N/A marks any cell lacking a validated execution run on that interface. For tasks requiring free-text open-ended dialogue evaluation, verifications use a standardized, frozen judge (\texttt{gpt-5.6-luna}) with rubric-grounded scoring prompts to guarantee comparability across vendors.

\subsection{Robustness to Execution and Verifier Choice}
\label{sec:robustness}

\paragraph{Conditional adherence and end-to-end success.}
Transient infrastructure errors (such as environment timeouts or interface communication anomalies) are distinguished from behavioral persona violations. On the evaluation set, \texttt{gpt-5-6-sol}'s chat full-pass rate is 68.2\% over 487 scored tasks, and 59.2\% when all 561 recorded chat attempts enter the denominator with non-completed trials conservatively scored as zero. For \texttt{grok-4-6} on app, the corresponding rate shifts from 69.9\% over 662 scored attempts to 67.5\% over 686 recorded attempts. Across all arms, assigning zero credit to incomplete attempts leaves relative model rankings and tier separations fully intact (\S\ref{sec:robustness}).

\paragraph{Model separation persists on deterministic checks.}
The benchmark declares 3,865 checks: 3,033 rule-based (78.5\%) and 832 judged. To verify that observed model separations do not rely on LLM judge behavior, we partition tasks by evaluator type and compare models only on shared scored tasks within each subset. For the GPT-6 pair, the full-pass gap is 17.7 points over 1,905 rule-only tasks and 11.4 points over 546 judge-only tasks. For the GPT-5.6 pair, it is 13.0 points over 1,891 rule-only tasks and 19.4 points over 474 judge-only tasks. The direction and magnitude of separation survive the complete removal of LLM judging, confirming that benchmark rankings reflect genuine model steerability rather than evaluator artifacts.

\paragraph{Failure-overlap cohort and case selection.}
The eighteen-arm chat/web intersection includes all model arms evaluated under the shared \texttt{gpt-5.6-luna} judge across both chat and web surfaces (1,086 common tasks; Figure~\ref{fig:cross-vendor}), fixing the persona profile, pinned traits, and rubric evaluator across models. Directed complementary-success counts are computed from binary full-pass outcomes without post-hoc independence assumptions. The qualitative case in Section~\ref{sec:competence-vs-fidelity} was inspected post hoc against the complete action trace, environment state, and persona specification to illustrate failure mechanisms where interface competence succeeds but latent constraints are violated.

\subsection{Repeat-Run Protocol and Empirical Stability}
\label{app:repeat-protocol}

\paragraph{Repeat sweeps.}
The stability study in Section~\ref{sec:stability} evaluates \texttt{gpt-6-sol} and \texttt{gpt-6-luna} across three independent evaluation passes per task under identical system prompts, environment specifications, and decoding parameters. Across surfaces, outcomes exhibit minimal run-to-run variation (standard deviations $\le 2.0$ points), confirming that leaderboard rankings reflect stable behavioral properties rather than sampling noise.

\paragraph{Stochasticity and agreement statistics.}
Pairwise repeat agreement is evaluated via Cohen's $\kappa$ across tasks completed across all three passes, pooling repeat pairs within each surface. The sol--luna score gaps and associated 95\% confidence intervals are estimated via 10,000 bootstrap task resamples, averaging task outcomes across repeats. As reported in Section~\ref{sec:stability}, interactive conversational turns (\env{chat}) exhibit expected natural dialogue stochasticity ($\kappa \approx 0.46$), whereas deterministic interface environments (\env{web} and \env{app}) yield high agreement ($\kappa \ge 0.63$), demonstrating that aggregate model comparisons remain highly reliable.

\subsection{Additional Four-Surface and Cross-Vendor Diagnostics}
\label{app:surface-diagnostics}

\begin{figure}[tbp]
\centering
\includegraphics[width=\linewidth]{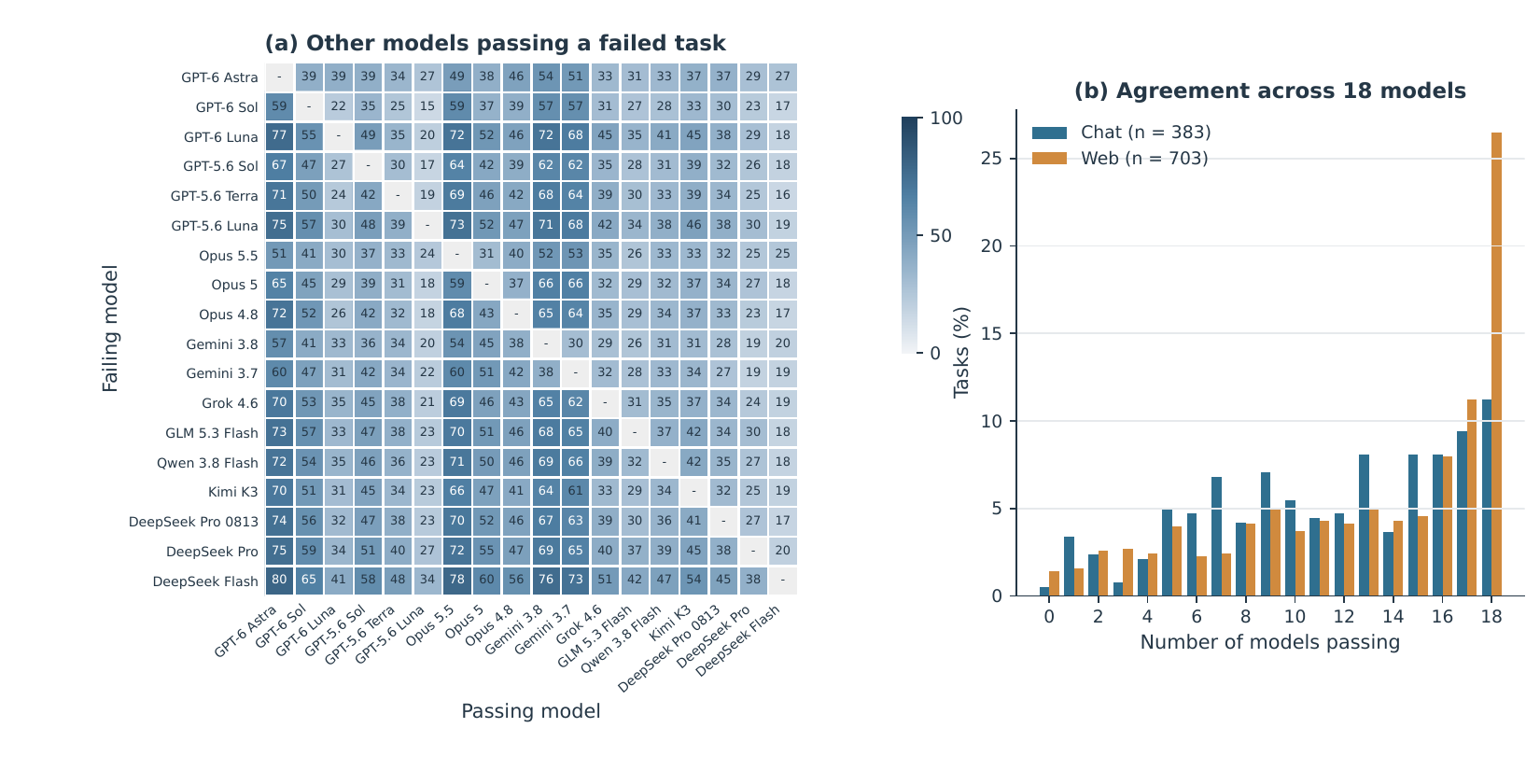}
\caption{\textbf{Task-level differences between models.}
(a) Percentage of a row model's failed tasks passed by the column model,
on 1,086 common chat/web tasks. Each row has its own failure denominator;
values are rounded to whole percentages. (b) Distribution of the number
of passing models, shown separately for chat and web. Both panels use
recorded full-pass outcomes with matching tasks, personas, and pinned traits.}
\label{fig:cross-vendor}
\end{figure}

The four-surface diagnostics evaluate the 140 matched single-attribute quartets under identical persona specifications. Let $Y_{m,t,s} \in \{0, 1\}$ indicate the full-pass verdict for model $m$, scenario $t$, and surface $s$. Conditional failure on surface $s$ given a pass on survey is computed as:
\[
\frac{\sum_t Y_{m,t,\mathrm{survey}}(1-Y_{m,t,s})}{\sum_t Y_{m,t,\mathrm{survey}}}.
\]
The total number of failed surfaces per scenario quartet is $4-\sum_s Y_{m,t,s}$. Pairwise cross-model complementarity on surface $s$ measures the proportion of model $m$'s failures that are successfully resolved by model $m'$:
\[
\frac{\sum_t (1-Y_{m,t,s})Y_{m',t,s}}{\sum_t (1-Y_{m,t,s})}.
\]
These descriptive metrics evaluate empirical complementary success across models without fitting parametric assumptions. The resulting cross-vendor pairwise agreement patterns across 1,086 common tasks are visualized in Figure~\ref{fig:cross-vendor}.

\section{Limitations}
\label{app:limitations}

We outline several methodological considerations and design trade-offs in \textsc{APB}, distinguishing foundational design choices from operational opportunities for future releases.

\paragraph{Construct validity and human fidelity.}
\textsc{APB} prioritizes \emph{internal construct validity} and causal attribution: tasks evaluate whether an agent acts consistently with declared latent traits under explicit, verified behavioral criteria, rather than predicting unconstrained human population distributions. By testing specific constraints without prompt leaks, the benchmark directly isolates model steerability from the confounding variance of human open-ended responses. While predicting real-world human behavioral distributions is an important complementary objective (which benefits from empirical telemetry of human users performing matching tasks), deterministic verification serves as a necessary prerequisite. Before evaluating whether an autonomous agent can emulate the nuanced distributional variance of human cohorts, one must first verify that its downstream actions reliably adhere to assigned profile constraints in everyday tasks.

\paragraph{Trait isolation, multi-attribute evaluation, and verifier precision.}
To maintain clear attribution, single-attribute tasks isolate individual traits within rich background profiles, treating other background attributes as distracting context. While real-world human behavior often involves subtle trait intersections, isolating specific constraints enables unambiguous error tracing. To assess combined demands, \textsc{APB} includes 1,172 multi-attribute tasks where multiple traits must hold simultaneously. While our primary metric requires full pass across all declared criteria to prevent partial-credit heuristics from masking constraint violations, we also report normalized credit across scored trials. Finally, for open-ended dialogue turns that rely on standardized LLM judges, we anchor evaluations to explicit rubrics and rely on deterministic rule-based predicates for the overwhelming majority of checks (78.5\%), with sensitivity audits confirming that model separations remain robust on deterministic checks alone (\S\ref{sec:robustness}).

\paragraph{Evaluation scale and empirical rank stability.}
Benchmarking 20 frontier model arms across 2,460 multi-turn tasks spanning four interaction surfaces requires executing over 45,000 full interactive episodes. To balance breadth across diverse model families and realistic desktop environments, the primary leaderboard reports a complete evaluation pass across all 20 arms. To verify that single-pass evaluations provide stable model comparisons, our repeated-sweep audit (\S\ref{sec:stability}) evaluates independent three-pass evaluations on representative arms, showing minimal run-to-run variation ($\sigma \le 2.0$ points). Furthermore, the 11 to 19 point separations between model tiers persist consistently across repeats, deterministic rule-based subsets, and error-inclusive analyses (\S\ref{sec:robustness}), confirming that aggregate leaderboard rankings capture robust architectural differences rather than runtime stochasticity.

\section{Future Directions and Validation Priorities}
\label{app:future-directions}

\paragraph{Log-grounded human realism: friction, hesitation, and abandonment.}
In realistic task environments, human behavior involves hesitation, partial disclosure, mid-dialogue correction, and task abandonment alongside compliant execution. Current benchmarks measure trait adherence as a conditioning response, testing whether an assigned characteristic appears in an interaction trace rather than whether the agent reflects authentic human dynamics. A primary direction is evaluating simulated trajectories against matched distributions of real user telemetry. At the population level, simulated turns can be benchmarked against production telemetry on turn length, phrasing variability, volunteered versus withheld context, correction rates, and abandonment frequencies. Distinguishing simulated traces from genuine human interactions with a statistical discriminator offers an interpretable summary metric, where an area under the ROC curve (AUC) near $0.5$ indicates well-calibrated simulation realism.

\paragraph{Individual-level predictive validation and causal trait intervention.}
At the individual level, validation requires evaluating whether persona models accurately predict how specific people act under novel tasks. This can be tested by extracting a persona profile from an initial conversation segment, prompting the model to simulate subsequent actions, and scoring simulated choices against historical actions under non-persona and shuffled-persona baselines. In parallel, experimental designs can progress beyond isolated single-trait testing toward counterbalanced factorial interventions ($2 \times 2$ designs). Systematically crossing orthogonal attributes (such as budget constraints with dietary preferences) within identical scenarios can isolate non-linear attribute interactions and quantify continuous adherence levels.

\paragraph{Long-horizon persona persistence and multi-agent collective dynamics.}
While \textsc{APB} tests models within bounded, single-session episodes, deployed persona agents often operate across extended horizons spanning days or weeks. Sustained deployment introduces risks of persona drift, forgetting of assigned constraints, and context collapse under accumulated dialogue history. A natural next step involves evaluating long-term memory architectures and parameter-efficient state tracking for continuous persona preservation. Finally, scaling persona evaluation to closed-loop multi-agent sandboxes, where heterogeneous persona agents interact in simulated marketplaces, organizational workflows, or negotiation games, will enable studying whether individual persona conditioning stably aggregates into plausible macro-level collective dynamics.

\clearpage
\section{Responsible Use and Artifact Release}
\label{app:responsible-use}

Simulated personas in \textsc{APB} are diagnostic instruments for evaluating model steerability and user simulation; they do not constitute statistical probability samples of real human populations or replace direct engagement with human stakeholders. Prohibited applications include impersonating named living individuals, generating surveillance or political targeting dossiers, scripting deceptive automated persuasion, or enforcing exclusionary discrimination against protected groups. To ensure reproducible and safe academic auditing, the 2,460 benchmark task definitions and persona attribute schemas are released under Creative Commons Attribution 4.0 International (CC-BY 4.0), while evaluation wrappers and execution harnesses are distributed under the Apache 2.0 license with cryptographic SHA-256 manifests. Practitioners applying \textsc{APB} to evaluate deployed applications should explicitly record the persona backbone model and cross-validate against heterogeneous model families to prevent shared-backbone evaluation bias.

\section{Ethical Considerations}
\label{app:ethics}

\paragraph{Synthetic personas, impersonation risks, and safety governance.}
\textsc{APB} simulates latent behavioral profiles to evaluate model steerability and consistency; it makes no normative claim that any real individual would or should behave as a declared persona indicates. A central ethical concern surrounding persona-conditioned models is the risk of deceptive anthropomorphism or malicious impersonation, wherein autonomous agents misrepresent their identity to manipulate users or mimic living individuals. We emphasize that \textsc{APB} does not endorse ungrounded impersonation in public deployments, where autonomous systems should consistently maintain clear, unambiguous disclosure of their artificial nature. Instead, empirical measurement is essential for assessing agentic systems: regulators, developers, and auditors cannot evaluate the behavioral boundaries of deployed agents without reproducible protocols that determine whether conditioning prompts govern real-world actions.

\paragraph{Stereotype mitigation, data privacy, and execution containment.}
Evaluating identity-linked attributes risks entrenching societal stereotypes if traits are naively bundled into caricatured personas. To mitigate this risk, \textsc{APB} decouples attributes through modular schema sampling, testing declared traits independently rather than relying on demographic archetypes. In addition, task scenarios focus on benign lifestyle, consumer, and professional settings, avoiding sensitive or high-stakes domains. All evaluated persona profiles are entirely synthetic, containing no personally identifiable information (PII) derived from real individuals. Finally, all interactive environments across web and native application surfaces are executed within isolated, reproducible sandboxes, ensuring that automated browser and desktop GUI operations produce no external network side effects, financial transactions, or interactions with external infrastructure.

\end{document}